\documentclass{article}

\PassOptionsToPackage{numbers, compress}{natbib}
 \usepackage[preprint]{neurips_2026}

\usepackage[utf8]{inputenc} % allow utf-8 input
\usepackage[T1]{fontenc}    % use 8-bit T1 fonts
\usepackage{hyperref}       % hyperlinks
\usepackage{url}            % simple URL typesetting
\usepackage{booktabs}       % professional-quality tables
\usepackage{amsfonts}       % blackboard math symbols
\usepackage{nicefrac}       % compact symbols for 1/2, etc.
\usepackage{microtype}      % microtypography
\usepackage{xcolor}         % colors
\usepackage{url}
\usepackage{multirow}
\usepackage{multicol}
\usepackage{float}
\usepackage{algorithm}
\usepackage{algorithmic}
\usepackage{float}
\usepackage{amsmath}
\usepackage{amssymb}
\usepackage{mathtools}
\usepackage{amsthm}
\usepackage{microtype}
\usepackage{graphicx}
\usepackage{subcaption}
\usepackage{booktabs}
\usepackage{wrapfig}
\usepackage{amsmath,amsfonts,bm}

\def\eqref#1{equation~\ref{#1}}
\def\1{\bm{1}}

\def\vx{{\bm{x}}}

\def\vz{{\bm{z}}}

\def\mI{{\bm{I}}}
\def\mJ{{\bm{J}}}

\DeclareMathAlphabet{\mathsfit}{\encodingdefault}{\sfdefault}{m}{sl}
\SetMathAlphabet{\mathsfit}{bold}{\encodingdefault}{\sfdefault}{bx}{n}

\def\sC{{\mathbb{C}}}

\DeclareMathOperator*{\argmax}{arg\,max}

\usepackage{etoc}
\usepackage{bbm}
\usepackage{tcolorbox}
\usepackage{tikz}

\addtocontents{toc}{\protect\setcounter{tocdepth}{-1}}

\title{Mitigating Diffusion Model Hallucinations with Dynamic Guidance}

\author{
Kostas Triaridis$^1$, Alexandros Graikos$^1$, Aggelina Chatziagapi$^1$,\\ \textbf{Grigorios G. Chrysos}$^2$\textbf{,} \textbf{Dimitris Samaras$^1$} \\\\
$^1$Stony Brook University, $^2$University of Wisconsin-Madison \\\\
\textbf{Project Page:} \url{https://cvlab-stonybrook.github.io/DynamicGuidance/}
\vspace{-20pt}
}

\begin{document}

\maketitle

\begin{abstract}
Hallucinations in diffusion models are samples with structural inconsistencies that can emerge due to the excessive smoothing of the learned score function, which in turn leads to interpolations between modes of the data distribution. Since semantic interpolations are often desirable and contribute to sample diversity, we believe that a nuanced and targeted solution is required to address diffusion model hallucinations. In this work, we introduce \textbf{Dynamic Guidance}, which \textit{mitigates} hallucinations by selectively sharpening the score function only along the pre-determined directions known to cause artifacts, while preserving valid semantic variations. This sharpening can be performed using either pre-determined classes or semantically coherent clusters that form pseudo-classes over the data distribution. The latter allows for a principled extension of Dynamic Guidance to text-to-image generation, where we select modes to correspond to fine-grained contextual differences in textual descriptions. To our knowledge, this is the first approach that addresses hallucinations at generation time rather than through post-hoc filtering. Dynamic Guidance substantially reduces hallucinations on both controlled and natural image datasets, significantly outperforming baselines. 
% Project page: \url{https://cvlab-stonybrook.github.io/DynamicGuidance/} \\
\end{abstract}

\section{Introduction}
Diffusion models have emerged as the dominant paradigm for image generation \citep{ho2020denoising, rombach2022high, chen2024pixart} due to their ability to generate high-fidelity and diverse images. Despite their success, they still remain prone to generating hallucinations \citep{aithal2024understanding}; i.e., samples that could never appear in the training set and are outside the support of the theoretical data distribution. A typical example in natural images are samples with incorrect anatomy, such as human hands with the incorrect number of fingers, or cats with missing body parts (See Figure~\ref{fig:t2i}).
Hallucinations have mostly been studied in the context of image-text misalignment, where generated images fail to sufficiently represent the information specified in a text prompt \cite{discovering_failures, H1, H2, li2025enhancing, zhao2025mitigating}.
In this work, we instead study hallucinations as a fundamental issue in the diffusion model's sampling process that can arise even when the generated samples adhere to the given conditioning.

\citet{aithal2024understanding} attributed hallucinations to mode interpolation, showing that models generate samples that lie between incompatible modes, i.e., regions of high probability density in the data distribution, producing semantically invalid content. They trace the mode interpolation issue to the fact that the learned score function of the model is excessively smooth between modes of the data distribution; the true score function is significantly sharper in the intermediate regions between modes, which means that the size of the required denoising steps in those regions is significantly larger. They verify that mode interpolation does not occur when using the true score on a mixture of Gaussians setting.

To avoid those low-probability regions, the denoising process must take larger steps, similar to the ones that would have been taken if the true score function had been used. Conventional guidance methods, such as classifier \citep{dhariwal2021cg} and classifier-free \citep{ho2021classifierfree} guidance, are designed to steer samples toward high-likelihood regions of the data distribution, typically to improve sample quality in conditional generation. They essentially sharpen the score function in directions that correspond to the given condition. Hallucinations arise in low-likelihood regions, motivating the use of guidance not only for improving fidelity but also for mitigating hallucinations during sampling. 

We argue that guidance with a pre-determined, fixed condition does not account for the full sampling trajectory; when the guidance condition and the initial noise are misaligned, the sample can be pushed into regions that require large corrective steps. The error in these steps scales with their magnitude (Figures~\ref{fig:norm_llama}, \ref{fig:norm_random}), which can cause the trajectory to overshoot or undershoot the target mode (Section~\ref{results}).

We also argue that we should be selective in reducing interpolations between modes \textbf{along the specific directions} where hallucinations occur. Some of the interpolations are welcome: they correspond to valid semantic variation and are essential for maintaining diversity in the generated samples (i.e., model generalization \citep{deschenaux2024going}), while others lead to implausible generations, perceived as hallucinations. For instance, in the latent manifold of hand images, interpolation along directions corresponding to skin tone yields valid and diverse samples, while interpolation along directions that control finger position may generate anatomically inconsistent hands with an incorrect number of fingers (Figure~\ref{fig:teaser}). We demonstrate that the score function can be sharpened \emph{selectively} by choosing guidance modes such that interpolations between them correspond precisely to hallucinations, thereby suppressing invalid generations without reducing diversity elsewhere (Section~\ref{score_sharpen}).

We propose \textbf{Dynamic Guidance} (DG), which adaptively selects the target for guidance at each denoising step. Instead of committing to a fixed condition from the start, a classifier is used to identify the most likely mode given the current state, and guidance is applied toward that mode. By allowing the target to change throughout sampling, Dynamic Guidance avoids being locked into trajectories that require large, interpolation-producing steps. Dynamic Guidance is the first method specifically designed to tackle hallucinations that mitigates hallucinations during the sampling process itself, rather than relying on post-hoc detection and rejection of flawed samples or post-training of the diffusion model. Crucially, by intervening directly in the generative process, our approach prevents hallucinations from arising in the first place. Mitigating hallucinations during sampling is preferable to post-hoc detection because it avoids wasting compute on samples that will later be discarded and preserves diversity by keeping the desired interpolations. We show that we can select the modes to correspond to pre-determined classes, like ImageNet-1k classes (Section~\ref{imagenet}), but that is not strictly necessary: Dynamic Guidance can work with pseudo-classes created by clustering semantically coherent neighborhoods in the data distribution (Section~\ref{sec:dino}). This allows for a principled extension of Dynamic Guidance to text-to-image generation, where we select modes to correspond to fine-grained contextual differences in textual descriptions (Section~\ref{sec:t2i}).

In summary, our contributions are as follows:
\begin{itemize}
    \item We propose Dynamic Guidance, a method that mitigates diffusion model hallucinations by adaptively guiding the denoising process away from low-probability regions.
    \item We use controlled experiments to demonstrate that Dynamic Guidance \textbf{selectively sharpens} the score function across directions that induce hallucinations, while preserving it elsewhere. 
    \item We present the first method to effectively mitigate hallucinations during the generation process instead of detecting them post-hoc. Our approach vastly outperforms previous detection-based methods in hallucination reduction on toy data and controlled datasets, achieving up to \textbf{70\% reduction in hallucination rate}.
    \item We apply Dynamic Guidance to 256 $\times$ 256 ImageNet generation and show that it improves hallucination-related metrics, consistently outperforming static guidance methods.
    \item We extend DG to text-to-image generation, with a user study showing that it consistently mitigates hallucinations, reducing the perceived hallucination rate for 100\% of participants.
\end{itemize}

\section{Dynamic Guidance}

Recent findings \cite{aithal2024understanding} have shown that the learned score function $s_{\theta}(\vx_t)$ in DDPMs tends to be overly smooth in the low-density regions between modes. This lack of ``sharpness" can effectively trap samples during denoising, a problem particularly pronounced by few-step sampling, where the denoiser cannot leap across these smooth regions. The resulting samples end up as interpolations between those modes, which might be ``hallucinations" depending on the semantic relationship between the specific modes involved.

We propose \textbf{Dynamic Guidance}, a simple yet effective guidance algorithm that adaptively sharpens the learned score function along the sampling trajectory. The core insight is to identify potential modes $c^*$ that a sample is naturally gravitating towards during the sampling process, and use the score function of the locally sharper conditional distribution $p(\vx\mid c^*)$ rather than the overly-smooth $p(\vx)$ or an attenuated $p(\vx\mid c)$ from a distant mode. 

However, the distribution modes $c$ are not always aligned with the conditioning $y$ a model has been trained with. In cases where the modes are more fine-grained than the conditions, hallucinations can often emerge as interpolations between such conditions. To solve this, we resort to inference-time guidance mechanisms with a post-hoc, finer set of conditions.

For our method, having first chosen appropriate mode labels $\sC$ whose interpolations correspond to the defined hallucinations (see Section~\ref{hallucination_datasets}: Single Shapes), we apply guidance dynamically, without fixing the guidance target/condition at the beginning of the sampling process. At each timestep, we identify the mode with the maximum probability given the current noisy sample $\vx_t$. and perform the sampling step by applying guidance using the selected mode. We essentially calculate a sharper approximation of the score function:
\begin{equation}
\hat s_{\theta}(\vx_t) := \sum_c \mathbbm{1}_{c^*}(c) \nabla_{\vx_t}\log p_\theta(\vx_t \mid c),
\quad c^* = \argmax_{c\in \sC} \log p(c | \vx_t).
\label{eq:dynamic_guidance}
\end{equation}
By recalculating the most probable class at each timestep, we ensure that the guidance signal remains aligned with the local score direction and adapts to the sample's evolving trajectory. For diffusion models with discrete conditioning, we pick class labels $\sC$ whose interpolations correspond to the defined hallucinations and calculate the score of $p(\vx|c^*)$ using classifier guidance \cite{dhariwal2021cg}. As an additional hyperparameter, we can control the strength of the guidance $\lambda$, effectively sampling using the score of the tempered distribution $p(\vx|c^*)^\lambda$. The proposed algorithm for DDIM sampling with DG using classifier guidance is summarized in Algorithm~\ref{alg:ddim}.
In Section~\ref{sec:t2i}, we show that the idea can be extended to models that use continuous conditions $c$, such as text tokens, and calculate the score of $p(\vx|c^*)$ using classifier-free guidance (CFG) \cite{ho2021classifierfree}.

\section{Dynamic Guidance selectively sharpens the score function}
\label{score_sharpen}

\begin{wrapfigure}[35]{r}{0.3\textwidth}
    \centering
    \includegraphics[width=1.0\linewidth]{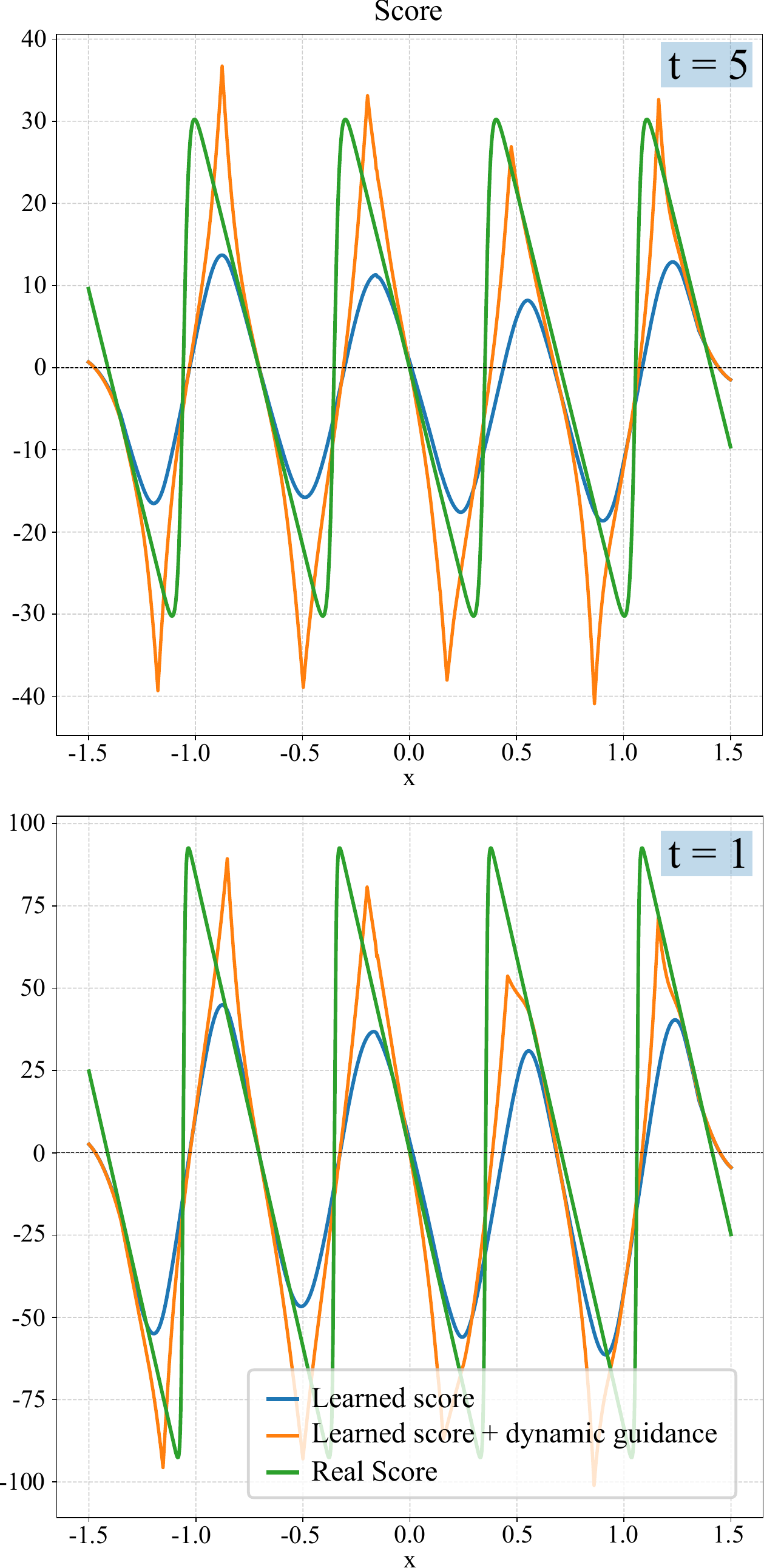}
    \caption{\textbf{Score Function Sharpening.} The learned score function of the diffusion model with and without Dynamic Guidance, compared to the true score function for a 2D mixture of Gaussians across the x dimension. The model learns a smoothed-out score function, which DG sharpens so that it more closely approximates the correct one.}
    \label{fig:score_gaussian}
\end{wrapfigure}

We find that sampling with Dynamic Guidance mitigates hallucinations because it selectively sharpens the score function learned by the diffusion model across the directions associated with the classifier. We first validate this in a simple 2D Gaussian setup, where the theoretical score function is given analytically. We calculate the unguided score function $s_\theta(\vx_t,t) = - \frac{\epsilon_{\theta}(\vx_t, t)}{\sqrt{1-\bar\alpha}}$ and the guided score function $\hat{s}_\theta(\vx_t,t) = s_\theta(\vx_t,t) + \lambda\nabla_{\vx_t}{\log p(c^*|\vx_t)}$ and plot them together with the real score function in Figure~\ref{fig:score_gaussian}. We see that Dynamic Guidance effectively sharpens the score function, which mitigates hallucinations as shown in Section~\ref{results}.

For more complex datasets, we want to visualize the learned score for specific directions that correspond to meaningful semantics. We use the Single Shapes dataset and examine the score function learned by a diffusion model when unguided and when using our proposed DG. In this dataset, all training images contain 1-3 instances of the \emph{same} shape, but diffusion models trained on it can generate images containing different shapes, which we consider hallucinations. Here, we focus on latent directions that correspond to changing the appearance of shapes, since variations there can lead to images containing different shapes, i.e hallucinations.

\begin{figure}[t]
  \centering
  % --- Left Side: Algorithm ---
\begin{minipage}[b]{0.37\textwidth}
\centering
    \includegraphics[width=0.95\linewidth]{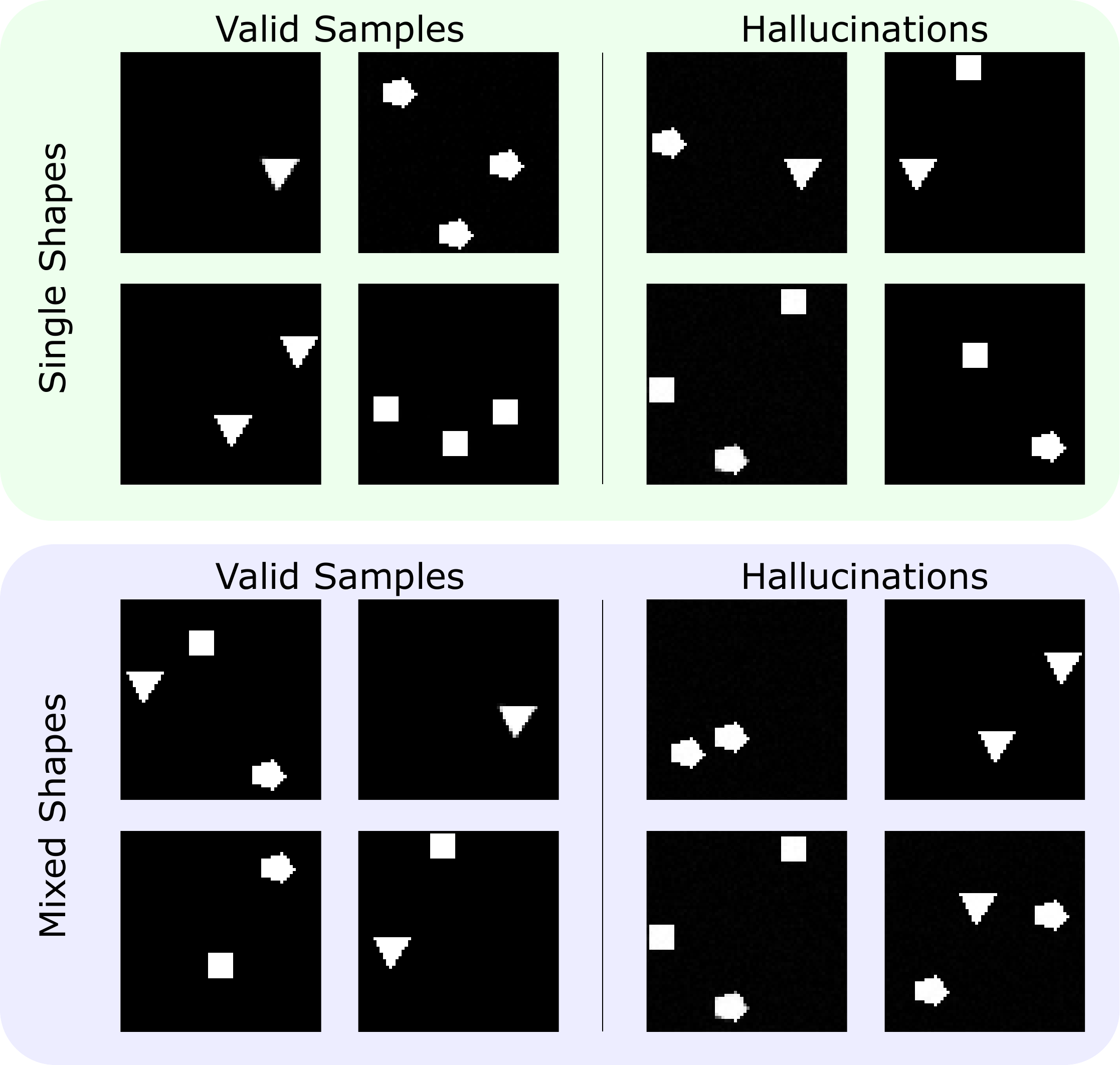}
    \subcaption[a]{}

    \label{fig:shapes}
\end{minipage}
\hfill
\begin{minipage}[b]{0.62\textwidth}
        \centering
    \includegraphics[width=\linewidth]{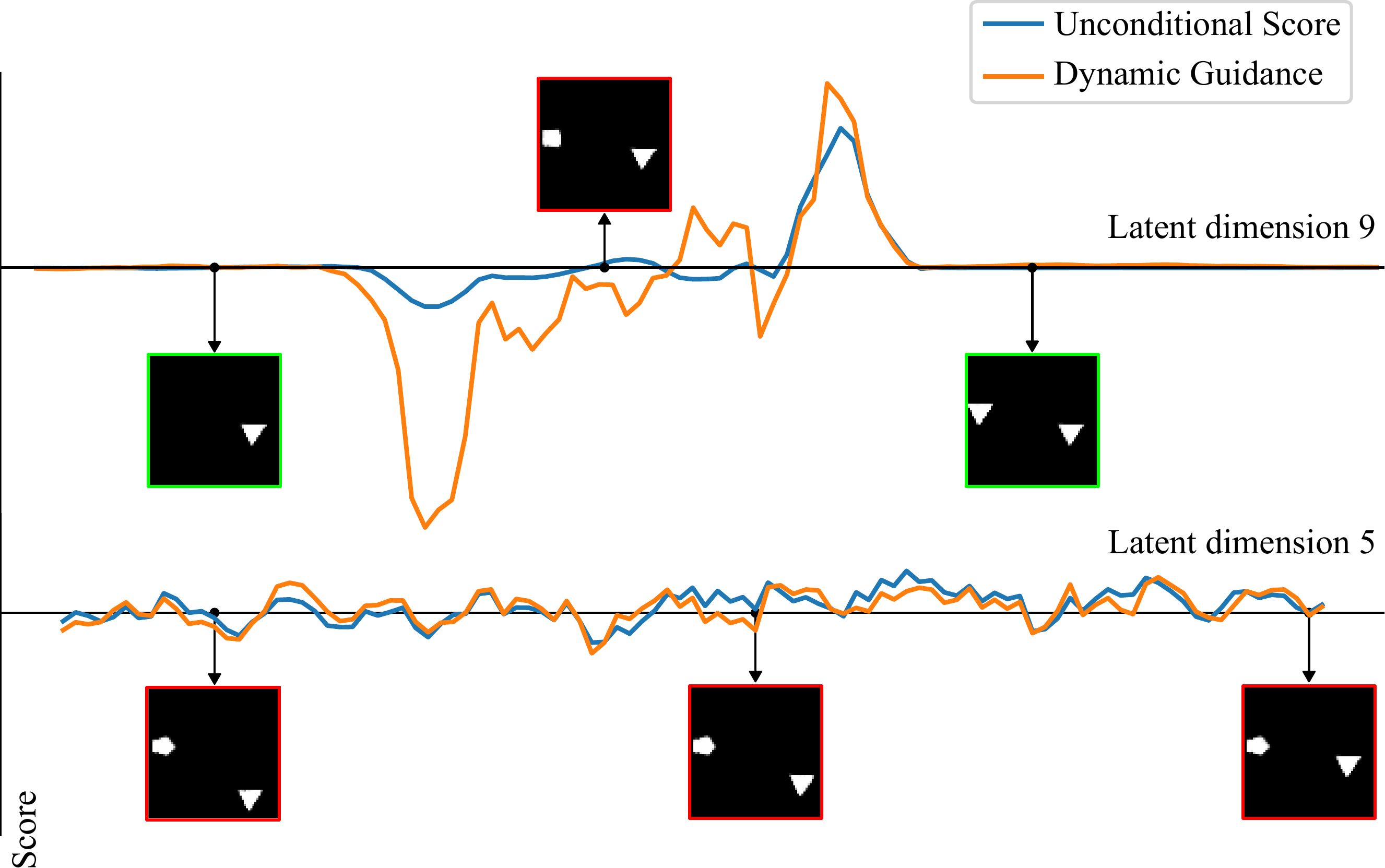}
        \subcaption[b]{}
    \label{fig:teaser}
    \end{minipage}
    \caption{\textbf{(a)} Examples of valid samples and hallucinations for the Single Shapes and Mixed Shapes datasets. \textbf{(b)} We pick an initial image that contains two different shapes (triangle + square), which is a hallucination for the Single Shapes dataset. We focus on a latent dimension that controls the appearance of the left shape \textbf{(Top)}. To resolve the hallucination, the square on the left should disappear or turn into a triangle. In the in-between region, where the left shape is square or pentagon, the unguided score function is zero, ``trapping" the sample and generating a hallucination. Dynamic Guidance sharpens the score in this region, steering the sample toward valid images that only contain triangles. Dynamic Guidance does not affect the score function along dimensions that are unrelated to hallucinations, like the one controlling the position of the shape on the right \textbf{(Bottom)}.}

\end{figure}

We first learn how to embed images from the dataset using a $\beta$-VAE \citep{higgins2017betavae,burgess2018betavae2} with a disentangled 10-dimensional latent representation. Examining the VAE-learned representations, we identify those that affect specific properties of the image: Dimension 9 controls the appearance of shapes on the left side, and Dimension 5 controls the position of shapes on the right (Figures~\ref{fig:teaser}, \ref{fig:latent_traversal}).

The trained $\beta$-VAE provides a transformation between latents and clean images using the learned encoder $\vz = \mathcal{E}(\vx_0)$ and decoder $\vx = \mathcal{D}(\vz)$. Our goal is to estimate the score of the latents $\nabla_{\vz} \log p(\vz)$. Using the change of variables formula for this autoencoder setting (\citep{cov}), we can express the distribution of the latents as
\begin{gather}
    p_{Z}(\vz) = p_{X_0}(\mathcal{D}(\vz))  \sqrt{\lvert\det \left(  \mJ_{\mathcal{D}}^T(\vz)\mJ_{\mathcal{D}}(\vz)  \right)\rvert},
    \label{eq:change_of_variables}
\end{gather}
where $\mJ_{\mathcal{D}}(\vz)$ is the Jacobian of the VAE decoder at $\vz$. Taking the $\nabla_{\vz} \log$ on both sides we have
\begin{gather}
    \begin{split}
    \nabla_{\vz} \log p_{Z}(\vz) &=\ \nabla_{\vz} \log p_{X_0}(\mathcal{D}(\vz)) 
      \\ &+ \nabla_{\vz} \log \sqrt{\lvert \det \left( \mJ_{\mathcal{D}}^T(\vz)\mJ_{\mathcal{D}}(\vz)  \right)\rvert}.
    \end{split}
    \label{eq:grad_log_z}
\end{gather}
To estimate the score of the latents $\nabla_{\vz} \log p_{Z}(\vz)$, we compute $\nabla_{\vz} \log \sqrt{\lvert\det \left(  \mJ_{\mathcal{D}}(\vz)^T\mJ_{\mathcal{D}}(\vz)  \right)\rvert}$ using automatic differentiation.
To calculate the score of the clean samples $\nabla_{\vz} \log p_{X_0}(\mathcal{D}(\vz))$ we first rewrite it as
\begin{gather}
    \nabla_{\vz} \log p_{X_0}(\mathcal{D}(\vz))
    = \mJ_{\mathcal{D}}^T(\vz) \nabla_{\vx_0} \log p_{X_0}(\vx_0),
    \label{eq:clean_score}
\end{gather}
and then we utilize the Perturb-and-Average Scoring from \citet{wang2023score}, which estimates the score of the clean images with an expectation over noisy image scores
\begin{gather}
    \nabla_{\vx_0} \log p_{X_0}(\vx_0) \approx \mathbb{E}_{\epsilon} 
    \left[ \nabla_{\vx_t} \log p_{X_t}(\sqrt{\alpha_t}\vx_0 + \sqrt{1-\alpha_t}\epsilon) \right].
    \label{eq:expectation_score}
\end{gather}
In practice, we first decode a latent $\vz$ into a clean image $\vx_0$. We then perturb $\vx_0$ with multiple random noise samples at timestep $t$ to get an approximation of the score of $\vz$ by averaging out the individual predicted scores.

To showcase how the score function is selectively sharpened, we select a generated image $\hat{\vx}_0$ that is a hallucination, showing a pentagon on the left and a triangle on the right. We plot the score function given by Equation (\ref{eq:grad_log_z}) along the latent dimension 9 (Figure~\ref{fig:teaser}). In this instance, the ideal score should be 0 in the regions that correspond to images with a single or two triangles. In contrast, the score for an image containing two shape types should be high, pushing the sample away from the hallucination.

Indeed, this is the exact observation we make in Figure~\ref{fig:teaser}; while the base diffusion-learned score does not push the sample away from the hallucination region, DG sharpens the score predictions, mitigating the hallucination. At the same time, when observing the score across a latent dimension that controls shape position, and thus is unrelated to hallucinations (latent dimension 5), we see that DG has no effect over it.

\section{Experiments}
\label{results}

\subsection{Controlled Settings}
\label{hallucination_datasets}

In this section, we describe the datasets used for our controlled study and define what is considered a hallucination for each. We also point to the labels used for the Dynamic Guidance.

\paragraph{Single Shapes:}
We construct a synthetic image dataset consisting of images containing triangles, squares, and pentagons. Each image contains one to three instances of the same shape, but never multiple shape types. In this setting, we define hallucinations as images containing different shapes simultaneously (see Figure~\ref{fig:shapes}). The class labels $c$ are given by the shape type, $c \in \{T,S,P\}$. Hallucinations, therefore, correspond to interpolations across these labels.

\paragraph{Mixed Shapes:}
We adopt the synthetic image dataset from \citet{aithal2024understanding} consisting of images of triangles, squares, and pentagons. Each image contains up to one instance of each shape, and may contain multiple shape types. In this setting, we define hallucinations as images containing 2 or more instances of the same shape (see Figure~\ref{fig:shapes}). The class labels $c$ are given by the combination of shapes, $c \in \{T,S,P, TS, TP, SP, TSP\}$. Hallucinations in this case do not necessarily correspond to interpolations across the selected class labels.

\paragraph{Hands:}
We use the Hands-11k dataset \citep{afifi201911kHands}, which contains images of human hands. Here, we define hallucinations as images that deviate from the expected hand anatomy.
For Dynamic Guidance, we use the 4 classes provided by the dataset, corresponding to the orientation of the hand: \textit{``palmar right''}, \textit{``palmar left''}, \textit{``dorsal right''}, and \textit{``dorsal left''}. Interestingly, in this setting, some hallucinations directly correspond to interpolations between the classes (images with hands facing down, with two thumbs are an interpolation between \textit{``dorsal right''} and \textit{``dorsal left''}), while others do not (5 fingers without a thumb). Dynamic guidance is effective in mitigating both.

\subsubsection{Results}

For all datasets we perform experiments for, we measure the reduction in hallucinations as follows:
\begin{equation}
    \text{HR} \;=\; 
    \frac{\;\#\text{Hallucinations}_{\text{before}} \;-\; \#\text{Hallucinations}_{\text{after}}}
         {\;\#\text{Hallucinations}_{\text{before}}}
\end{equation}

\begin{table*}[h]
\centering
\small

\begin{minipage}{0.32\linewidth}
\centering
\caption{Single Shapes dataset.}
\label{table:single_shapes}
\begin{tabular}{lc}
\toprule
\multicolumn{1}{c}{{Method}} & HR$\uparrow$\\
\midrule
Var. Filtering (1\%)                      & 1.05\%  \\
Var. Filtering (5\%)                    & 3.16\%     \\
Var. Filtering (10\%)              & 11.58\%     \\
Classifier Guidance          &     11.05\%  \\
\textbf{Dynamic Guidance}          & \textbf{74.21\%} \\
\bottomrule
\end{tabular}
\end{minipage}
\hfill
\begin{minipage}{0.32\linewidth}
\centering
\caption{Mixed Shapes dataset.}
\label{table:mixed_shapes}
\begin{tabular}{lc}
\toprule
\multicolumn{1}{c}{{Method}} & HR$\uparrow$ \\
\midrule
Var. Filtering (1\%)                      & 1.35\%       \\
Var. Filtering (5\%)            & 8.11\%                  \\
Var. Filtering (10\%)              & 17.57\%        \\
Classifier Guidance          &     -1.92\%  \\
\textbf{Dynamic Guidance}          & \textbf{72.97\%}   \\
\bottomrule
\end{tabular}
\end{minipage}
\hfill
\begin{minipage}{0.32\linewidth}
\caption{Hands-11k dataset.}
\label{table:hands}
\begin{tabular}{lc}
\toprule
\multicolumn{1}{c}{{Method}} & HR$\uparrow$ \\
\midrule
VF (1\%)                      & 1.83 ± 3.68\%      \\
VF (5\%)            & 5.46 ± 5.16\%                 \\
VF (10\%)              & 9.19 ± 5.99\%        \\
CG          &     17.81 ± 17.88\%  \\
\textbf{DG}          & \textbf{45.12 ± 5.36\%}   \\
\bottomrule
\end{tabular}
\end{minipage}
\end{table*}

Negative values correspond to an increase in hallucinations. For all controlled image datasets, we train an ADM \citep{dhariwal2021cg} and a noisy sample classifier using the guided-diffusion\footnote{https://github.com/openai/guided-diffusion} codebase. We compare to variance filtering, the detection-based method proposed by \cite{aithal2024understanding}, and Classifier Guidance. For variance filtering, we pick the best hyperparameters by performing a grid search around the values mentioned in their paper, and evaluate using different discard rates. Across these diverse benchmarks, Dynamic Guidance (DG) consistently outperforms variance filtering (VF) and Classifier Guidance (CG) in mitigating hallucinations, achieving over a 70\% reduction in the Single and Mixed Shapes datasets (Tables \ref{table:single_shapes}, \ref{table:mixed_shapes}) and over 40\% in the Hands-11k dataset with 25-step DDIM sampling.

\subsection{ImageNet}
\label{imagenet}

Hallucinations in large-scale image benchmarks span a vast range of classes, objects, and scenes, making them nearly impossible to precisely define, let alone detect.
With this in mind, we choose ImageNet1k generation as a large-scale benchmark to evaluate our approach. In this setting, hallucinations cannot be strictly defined, so we rely on proxy metrics: precision, recall \citep{precisionrecall}, and Inception Score \citep{inceptionscore}. Precision measures the fraction of generated samples that fall inside the support of the estimated real data distribution. We argue that higher precision reflects fewer hallucinations, since hallucinations correspond to samples outside the \textit{true} data distribution. Inception Score is maximized when generated images clearly belong to some class (they are low-entropy, high-confidence predictions for the classifier) and diverse (labels evenly distributed). When generating samples, conditioning and initial noise are often mismatched \citep{li2025enhancing} and images fail to correspond to any valid class, visually resembling hallucinations (Figure~\ref{fig:imagenet}). A large presence of such samples reduces the classifier’s confidence and consequently hurts both the Inception Score and Precision.

\begin{figure*}[!t]
    \centering
    \includegraphics[width=\linewidth]{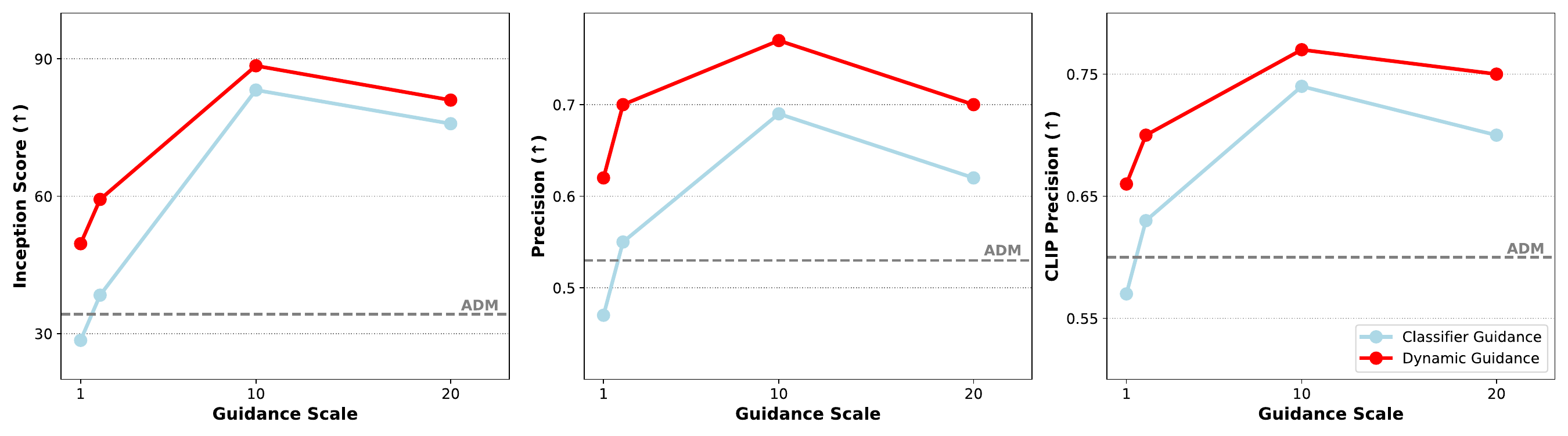}
    \caption{Performance of Classifier and Dynamic Guidance on Hallucination-related metrics on ImageNet-1k generation. Dynamic Guidance consistently improves Inception Score and precision across guidance scales, which corresponds to fewer hallucinations.}
    \label{fig:imagenet_results}
\end{figure*}

In our experiments, we evaluate precision and recall using both the Inception \citep{inception} and CLIP \citep{clip} models and Inception Score. We adopt the pretrained diffusion model and classifier from \citet{dhariwal2021cg} and compare three settings: unguided generation, Classifier Guidance (CG) with varying guidance scales, and Dynamic Guidance (DG) with the same scales. The results, summarized in Table~\ref{table:imagenet}, show that Dynamic Guidance consistently outperforms Classifier Guidance across all settings (Figure~\ref{fig:imagenet_results}). Notably, in the best setting, it achieves precision and Inception Score gains of 8 and 5 points, respectively, while maintaining diversity; recall is not significantly reduced, and the Inception Score is the highest overall. This trend holds across different metrics; we also measure FID with both Inception and CLIP and generative density and coverage \citep{naeem2020reliable} and report the results in Tables~\ref{fid_table} and~\ref{table:imagenet_density_coverage} of Appendix Section~\ref{fid}. In Appendix Figures~\ref{fig:norm_llama}, \ref{fig:norm_random}, we show how `static' CG tends to overshoot in cases where the selected label does not align with the initial noise sample.

\subsubsection{Selection of guidance interval}
We identify that, unlike Classifier Guidance, our proposed Dynamic Guidance (DG) can be performed just for a subset of timesteps, which we denote in our algorithm as [$T_1,T_2$]. Classifier Guidance attempts to impose a \textbf{strong constraint} on the generation process: the final sample should belong to the chosen class. On the other hand, Dynamic Guidance is more similar to Classifer-Free Guidance (CFG) in that it attempts to guide a strong signal (conditional model in the case of CFG, unconditional model in the case of DG) with a weak guidance signal (unconditional model in the case of CFG, dynamic gradient of a classifier in the case of DG) to improve generation.

Inspired by recent work on CFG \citep{kynkaanniemi2024applying, wang2024analysis}, we show that applying DG only for some intermediate timesteps [$T_1,T_2$] improves performance. To select $T_1$ and $T_2$ for each experiment, we chose the generation timestep at which the image begins to form ($T_1$), and the timestep where samples appear to have already converged to an image that cannot be modified further ($T_2$). We verify our choice of $[T_1,T_2]$ in Table~\ref{table:t1t2}, where for our ImageNet experiments the selected $T_1=800$ and $T_2=400$ achieve the best results. In Figures \ref{fig:bisonvllama} and \ref{fig:echidna} in the Appendix, we visualize the generation process to show how our choice of $T_1$ and $T_2$ is motivated by the image formation process.

\begin{table*}[h!]
\centering
\small

\begin{minipage}[t]{0.48\linewidth}
\centering
    \small
    \caption{Ablation for the timestep interval [$T_1,T_2$] used to perform Dynamic Guidance in ImageNet.}
    \label{table:t1t2}
    \begin{tabular}{lccc}
    \toprule
    Method & IS$\uparrow$ & Prec$\uparrow$ & Rec$\uparrow$ \\
    \midrule
    Uncond. ADM          & 34.24 & 0.53 & 0.61 \\
    \midrule
    + DG [1000-0]   & 48.20&	0.73&	0.33 \\
    + DG [600-200]  & 61.07&	0.64&	\textbf{0.62} \\
    \textbf{+ DG [800-400]}  & \textbf{88.49}&	\textbf{0.77}&	0.52 \\
    \bottomrule
    \end{tabular}
\end{minipage}
\hfill
\begin{minipage}[t]{0.48\linewidth}
    \centering
    \small
    \caption{Comparison of Dynamic Guidance with a classifier trained on DINOv2 pseudo-classes to one trained on real ImageNet classes.}
    \label{table:dino}
\begin{tabular}{l@{\hspace{0.6em}}c@{\hspace{0.5em}}c@{\hspace{0.5em}}c@{\hspace{0.5em}}c@{\hspace{0.5em}}c@{\hspace{0.5em}}}
\toprule
&  \multicolumn{3}{c}{Inception} & \multicolumn{2}{c}{CLIP} \\
\cmidrule(lr){2-4} \cmidrule(lr){5-6}
Method & IS$\uparrow$ & Prec$\uparrow$ & Rec$\uparrow$ & Prec$\uparrow$ & Rec$\uparrow$ \\
\midrule
Uncond. ADM          & 34.24 & 0.53 & 0.61 & 0.60 & 0.26 \\
\midrule
+ CG w/ real  & 83.19 & 0.69 & 0.55 & 0.74 & 0.27 \\
\midrule
+ DG w/ real & 88.49 & 0.77 & 0.52 & 0.77 & 0.26 \\
\textbf{+ DG w/ clusters} & 75.17 & 0.78 & 0.51 & 0.77 & 0.25 \\
\bottomrule
\end{tabular}
\end{minipage}
\end{table*}

\subsubsection{Guidance with Pseudo-Classes}
\label{sec:dino}
In the Mixed Shapes experiments, we discussed how DG can be used with a classifier whose labels do not directly correspond to the hallucinations we aim to avoid. The set of labels we use for the purpose of reducing hallucinations with DG does not have to be identical to the set of labels used to control a generative model. The labels used for conditioning must correspond to human-interpretable semantics, such as object categories, attributes, or text, since they determine what the model is intended to generate. In contrast, the labels used for hallucination reduction do not need to carry such semantic meaning; they may correspond to abstract latent modes or auxiliary partitions of the data that help isolate hallucination-prone directions without mapping to interpretable concepts.

For our ImageNet experiments, we create pseudo-classes by clustering the training images using DINOv2 \citep{oquab2024dinov2}. We create 5000 clusters using the hierarchical clustering method described by \citet{vo2024automatic}, assign pseudo-labels to the clusters, and train a classifier to predict those pseudo-labels. In Table~\ref{table:dino}, 
Dynamic Guidance with a classifier trained on DINOv2 pseudo-classes performs as well as DG with ImageNet classes in terms of precision and recall. With regard to InceptionScore DG with pseudo-classes still improves performance compared to the unguided model, but underperforms DG with ImageNet classes. We attribute this to the bias of IS towards exact ImageNet classes specifically.

In the Appendix Figure~\ref{fig:cond} we also show that it is possible to use distinct sets of labels for conditioning (ImageNet classes) and guiding a model with DG (DINO clusters). The two can act orthogonally and DG can improve generations even in the class-conditional case. This means that it is possible to pick labels that correspond to required semantics for training a conditional model and a distinct set of labels that help reduce hallucinations when used with DG.

\subsection{Text-to-Image generation}
\label{sec:t2i}
\begin{figure}[t!]
    \centering
    \includegraphics[width=\linewidth]{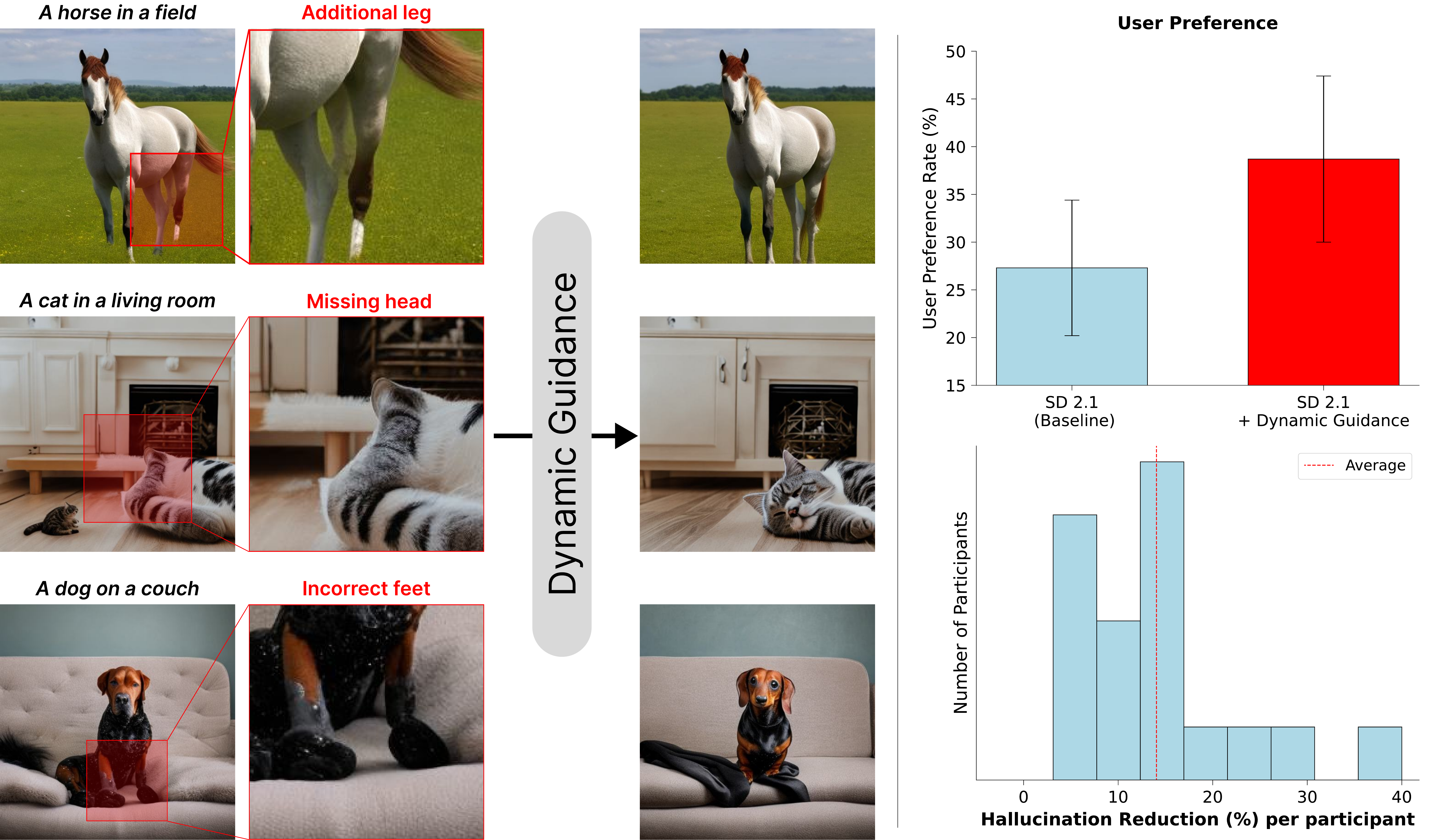}
    \caption{(\textbf{Left}) Text-to-image models can generate hallucinated images that contain anatomical errors, such as extra legs, or a missing head. Dynamic Guidance effectively mitigates those hallucinations, producing higher-quality samples. (\textbf{Right}) We evaluate Dynamic Guidance on text-to-image generation with a user study. Remarkably, Dynamic Guidance reduces the hallucination rate for \textbf{all} participants, with an average reduction in hallucinations of 14.1\%. Participants also showed strong preference for images generated using Dynamic Guidance over the baseline.}
    \label{fig:t2i}
\end{figure}

We extend Dynamic Guidance to text-conditioned latent diffusion models \cite{rombach2022high}. In text-to-image models, we identify that an ambiguous prompt, such as \textit{``a horse in a field''} leaves the horse's pose and action unspecified, which regularly causes hallucinations. During denoising, the model may hedge among multiple valid interpretations, resulting in anatomical errors (e.g., malformed limbs or improbable poses).

To address this, we propose an approximation to the sharpening of the score function by modifying the prompt during sampling. Analogous to the $\argmax$ operation on a classifier, we employ a Vision-Language Model (VLM) to observe intermediate denoised predictions and dynamically adjust the text prompt during generation. 
We utilize a set of classes designed to resolve ambiguities that cause hallucinations; given a specific input prompt $P$, we ask a Large Language Model (LLM) to identify the primary axes of ambiguities contained in the prompt. Then the LLM is tasked with creating a diverse set of branches, designed to resolve this ambiguity: for example, for the \textit{``a horse in a field''} prompt, the LLM proposes branches to resolve the pose and action ambiguity such as \textit{``a horse galloping with legs extended in a field''} and \textit{''a horse lying down in a field''}. 

During denoising, we allow the text conditioning to adapt within a guidance window $[T_{\text{1}}, T_{\text{2}}]$. For each step, we calculate the current clean latent prediction $\vz_0(\vz_t)$, decode it into an image using the VAE, and ask a VLM to identify the closest match given multiple options corresponding to the created branches. If the VLM identifies a match, we update the text prompt and continue sampling; otherwise, the VLM may also respond ``STAY'' if the image is too noisy to classify, in which case we retain the current prompt. We show that Guidance with a dynamically changing prompt can effectively mitigate hallucinations, especially the ones corresponding to incorrect anatomy (Figure~\ref{fig:t2i}).

We implement Dynamic Guidance on top of Stable Diffusion 2.1 \cite{rombach2022high}, and use a classifier-free guidance scale of 7.5 \cite{ho2021classifierfree} for sampling. We use Claude \cite{anthropic2024claude3} to create the branches that are used as ``classes'' and Qwen2-VL-7B \cite{wang2024qwen2} as the classifier. We evaluate Dynamic Guidance against the Stable Diffusion 2.1 baseline with a user study of 18 participants. To highlight the simplicity of direct application of Dynamic Guidance on a given text-to-image model we choose some reasonable default hyper-parameters, avoiding the need for extensive tuning (More details in Appendix Section~\ref{sec:impl_t2i}). We recognize that the definition of hallucinations in complex image generation settings is quite ambiguous, and identify that, even when given strict guidelines, different people have different ``thresholds'' for identifying a sample as a hallucination. Even with this variance, \textbf{Dynamic Guidance remarkably reduced the perceived hallucination rate for all participants}. In total, Dynamic Guidance reduced the perceived hallucination rate for 70\% of prompts, resulting in an average reduction of 14.1\%, from 52.7\% to 45.3\%. Additionally, participants preferred images generated using Dynamic Guidance with a rate of 38.7\% versus 27.3\% for the baseline. We also show that using the generated, more complex, branches as base prompts can lead to worse generations and new hallucinations (similarly to the observations by \citet{park2025raretofrequent}). We show examples in which those are resolved by guiding adaptively, using our proposed Dynamic Guidance in Figure~\ref{fig:t2i_prompt}.

\section{Related Work}
\paragraph{Diffusion Models} Introduced by \citet{sohl2015deep}, diffusion models \citep{ho2020denoising, song2020score} 
are characterized by the forward process, where Gaussian noise is gradually added to a sample, and a learned reverse process, where a network learns to denoise samples. The original denoising diffusion formulation was shown to be equivalent to score-matching \citep{hyvarinen2005estimation, vincent2011connection,song2020score}, linking the denoiser predictions to the score function of the noisy data. We adopt the score-function view of diffusion models in this paper to study the generation and mitigation of \textit{hallucinations}. 

\paragraph{Guiding Diffusion Models} The reverse diffusion process can be controlled to draw samples with constraints \citep{dhariwal2021cg, ho2021classifierfree, graikos2022diffusion}. \citet{dhariwal2021cg} introduced an external classifier into the diffusion model's sampling, using its gradients to influence the sampling trajectory towards a target class, while \citet{ho2021classifierfree} jointly trained a conditional and an unconditional diffusion model and sampled from a combination of their scores. A more recent work \citep{karras2024guiding} suggests that it is possible to improve the fidelity of the generations by guiding the model with a weaker one. While those guidance methods have aimed to improve the fidelity of generations by strategically sampling from well-learned high probability regions, another line of work has focused on incorporating arbitrary training-free guidance by imposing specific linear \cite{chung2023dps} or non-linear \cite{yu2023freedom} constraints, either through backpropagation \cite{chung2023dps, yu2023freedom} or approximate Newton iterations \cite{graikos2025fast}.

\paragraph{Hallucinations in Diffusion Models} Prior work has examined hallucinations in diffusion models \citep{aithal2024understanding, H1, H2}, but these are most often defined in terms of text–image misalignment \citep{li2025enhancing,discovering_failures, discovering_failures2}, where the diffusion model fails to generate an image that matches the conditioning prompt. Such hallucinations are more linked to issues of text-image alignment and text representations, and most approaches address them by improving prompt adherence through post-training \citep{pt1, pt2, pt3}. In contrast, we take a view of hallucinations that is more directed to the diffusion model: rather than treating them as failures of the text condition and the model's adherence to it, we study them as a fundamental property of the diffusion model itself. Our focus is on the model’s score function and its tendency to interpolate across modes in ways that yield unrealistic samples, independent of the text condition.

\paragraph{Mode Interpolation} \citet{aithal2024understanding} analyze the problem of hallucinations in diffusion models through mode interpolation. This, however, is not the first time the phenomenon of mode interpolation has been observed in generative models, and not all instances of mode interpolation result in hallucinations. \citet{deschenaux2024going} show that you can guide a diffusion model to produce interpolations of \emph{desired} attributes not present in the training data, while there are other recent works that detect ``mode mixture'' in GANs \citep{An2020AE-OT:} and diffusion models \citep{li2023dpm}.

\section{Limitations}
While Dynamic Guidance effectively mitigates hallucinations, it can also impact the diversity of generated samples, as it also introduces certain biases. The impact of Dynamic Guidance on the diversity-hallucination trade-off depends on how the selected classes relate to the hallucination direction. When the chosen classes align with hallucination-relevant directions (e.g., Single Shapes), Dynamic Guidance sharpens the score only where necessary, preserving diversity. When this alignment is not possible, it still mitigates hallucinations but may slightly reduce diversity, as suggested by the modest recall drop and increased coverage on ImageNet (Section~\ref{fid}). Since the method relies on a classifier to determine the most likely mode at each timestep, any bias in the classifier’s predictions can affect the sampling trajectory, leading to a preference for conditions that are easier to identify. Consequently, certain semantic modes can receive disproportionately more probability mass, leading to biased final distributions. In addition, the initial noise introduces bias in the composition of the generated image, further skewing the distribution of generated samples. In practice, we find that certain classes are over-represented (Figures \ref{fig:gen_classes_1} and \ref{fig:gen_classes_10}) when using DG.

\section{Conclusion}
In this work, we addressed the problem of hallucinations in diffusion models by introducing Dynamic Guidance (DG), a method that mitigates hallucinations during the generative process itself. Unlike prior detection-based approaches, DG prevents hallucinations from arising by selectively sharpening the score function along hallucination-inducing directions while preserving benign interpolations that support diversity. Our experiments across toy data, controlled and real-world image datasets demonstrate consistent improvements, with DG achieving substantial hallucination reduction even under realistic low-step DDIM sampling. On the large-scale benchmark ImageNet, we further show improvements in proxy metrics such as precision and Inception Score, validating that our method generates samples that remain closer to the true data distribution. Finally, we showed that DG can also be applied to text-to-image models, effectively mitigating hallucinations. We believe Dynamic Guidance provides a principled step toward more reliable diffusion models and opens opportunities for future work in understanding and controlling hallucinations in large-scale generative models.

\section*{Acknowledgements}
This research was supported by NSF grants IIS-2123920, IIS-2212046.

\bibliographystyle{abbrvnat}
\bibliography{main}

%%%%%%%%%%%%%%%%%%%%%%%%%%%%%%%%%%%%%%%%%%%%%%%%%%%%%%%%%%%%
\newpage
\appendix
\section*{Appendix}
\addtocontents{toc}{\protect\setcounter{tocdepth}{2}}
\tableofcontents

\clearpage

\section{Background on Diffusion Models}
\subsection{Denoising Diffusion Probabilistic Models} DDPMs \citep{ho2020denoising} learn to draw samples from a given data distribution $q(\vx)$. They consist of a forward process, in which Gaussian noise of increasing variance, controlled by a pre-defined schedule $\bar{\alpha}_t$, is iteratively added to a sample $\vx_0 \sim q(\vx)$ to produce a noisy sample $\vx_t$, and a reverse process that learns how to denoise samples by predicting the added noise with a neural network $\epsilon_{\theta}(\vx_t, t)$.

Once the denoising network is trained, DDPMs can draw new samples, starting from random Gaussian noise $\vx_T \sim \mathcal{N}(\mathbf{0}, \mI)$, by following the inverse process transitions
\begin{align}
    p_\theta(\vx_{t-1}|\vx_t) = \mathcal{N}(\vx_{t-1}; \mu_\theta(\vx_t, t), \Sigma_\theta(\vx_t, t)),
\end{align}
where the mean is given from the predicted noise $\epsilon_{\theta}(\vx_t, t)$ and the variance $\Sigma_\theta(\vx_t, t))$ can either be fixed or learned \citep{nichol2021improved, dhariwal2021cg}.

\subsection{Denoising Diffusion Implicit Models} \citet{song2020ddim} introduced DDIM, which allows deterministic sampling from a trained diffusion model with fewer steps. We adopt this sampling approach when using fewer than 100 sampling steps, following \citet{nichol2021improved}. In DDIM, the reverse sampling steps are defined as:
\begin{equation}
x_{t-1} = 
\sqrt{\bar{\alpha}_{t-1}} \underbrace{ 
\left( \frac{x_t - \sqrt{1-\bar{\alpha}_t}\,\epsilon_\theta(x_t, t)}{\sqrt{\bar{\alpha}_t}} \right) }_{\text{``predicted $\vx_0$''}} 
+  \sqrt{1-\bar{\alpha}_{t-1}} \underbrace{ \,\epsilon_\theta(x_t, t) }_{\text{``direction pointing to $x_t$''}} ,
\end{equation}
where the model uses a \textit{prediction} of the clean image to perform the denoising.

\subsection{Connections to Score Based Generative Models} The score function \( s(x) \) of a probability distribution \( p(x) \) is defined as the gradient of the log probability density function, \(s(x)= \nabla_x \log p(x) \). Score-based generative modeling \citep{song2019generative} aims to learn this score function of the data distribution from samples drawn from the same distribution.

In the context of diffusion models, denoising diffusion has been shown to also approximate the score function \citep{song2020score}
\begin{equation}
s_{\theta}(x_t, t) = -\frac{\epsilon_\theta(\vx_t, t)}{\sqrt{1 - \bar{\alpha}_t}}.
\end{equation}

\subsection{Classifier Guidance}
Data generated by diffusion models often fail to reproduce the clarity of the training data. A widely-used technique to increase fidelity of samples is classifier guidance \citep{dhariwal2021cg}, which uses the gradient of a classifier $p(y|\vx_t)$, trained on noisy samples $\vx_t$, to guide the denoiser network towards synthesizing more realistic samples. Classifier guidance modifies the predicted noise from a network with a term that maximizes the classifier likelihood
\begin{align}
\epsilon_{\theta}'(\vx_t, t) = \epsilon_{\theta}(\vx_t, t) - \lambda \sqrt{1-\bar{\alpha}_t} \nabla_{\vx_t} \log p(y \mid \vx_t) ,
\end{align}
where $\lambda$ is a hyperparameter, controlling the strength of the guidance. 

\section{Implementation Details}
\subsection{Dynamic Guidance Algorithm with DDPM, DDIM and Classifier Guidance}
We describe the algorithm using DDPM in Algorithm~\ref{alg:ddpm_dg}. In the main paper, we employed DDIM (Algorithm~\ref{alg:ddim}) as it is the more practical sampling algorithm for models larger than the toy 2D Gaussian setting. 

\label{app_ddpm}

\begin{center}
\begin{minipage}{.48\linewidth}
    \begin{algorithm}[H]
    \centering
    \caption{Dynamic Guidance with DDPM}
    \small
    \begin{algorithmic}
    \STATE \textbf{Input:} timesteps $T$, dynamic guidance steps $T_1$ $T_2$, denoiser $\mu\theta$, classifier $p_{\phi}$
    \STATE $\vx_T \sim \mathcal{N}(0, \mathbf{I})$
    \FOR{$t = T \dots 1$}
        \STATE $\mathbf{z} \sim \mathcal{N}(0, \mathbf{I})$
        \IF{$T_1\ge t \ge T_2$}
        \STATE $y^* = \argmax p_{\phi}(y | \vx_t)$
        \STATE $\vx_{t-1} = \mu_\theta(\vx_t, t) + \sigma_t^2 \lambda \nabla_{\vx_t} \log p_{\phi}(y^* | \vx_t) + \sigma_t \mathbf{z}$
        \ELSE
        \STATE $\vx_{t-1} = \mu_\theta(\vx_t, t) + \sigma_t \mathbf{z}$
        \ENDIF
    \ENDFOR
    \STATE {\bfseries Return} $\vx_0$
    \end{algorithmic}
    \label{alg:ddpm_dg}
    \end{algorithm}
\end{minipage}
\hfill
\begin{minipage}{0.48\linewidth}
\begin{algorithm}[H]
\caption{Dynamic Guidance with DDIM}
\small
\label{alg:ddim}
\begin{algorithmic}
\STATE \textbf{Input:} timesteps $T$, dynamic guidance steps $T_1$ $T_2$, denoiser $\epsilon_\theta$, classifier $p_{\phi}$
\STATE $\vx_T \sim \mathcal{N}(0, \mI)$
\FOR{$t = T \dots 1$}
    \IF{$T_1\ge t \ge T_2$}
    \STATE $y^* = \argmax_y \log p_{\phi}(y | \vx_t)$
    \STATE $\hat\epsilon = \epsilon_\theta(\vx_t) - \lambda\sqrt{1-\bar{\alpha}} \nabla_{\vx_t} \log p_{\phi}(y^* | \vx_t)$
    \ELSE
    \STATE  $\hat\epsilon = \epsilon_\theta(\vx_t)$
    \ENDIF
    \STATE $\tilde{\vx}_0  = \frac{1}{\sqrt{\bar{\alpha}_t}} \left( \vx_t - \sqrt{1 - \bar{\alpha}_t} \hat\epsilon_\theta(\vx_t, t) \right)$
    \STATE $\vx_{t-1}  = 
\sqrt{\bar\alpha_{t-1}} \tilde{\vx}_0
+  \sqrt{1-\alpha_{t-1}}\,\hat\epsilon_\theta(\vx_t, t)$
\ENDFOR
\STATE {\bfseries Return} $\vx_0$
\end{algorithmic}
\end{algorithm}
\end{minipage}

\end{center}

\subsection{2D Mixture of Gaussians}
We create a synthetic toy dataset with a mixture of 25 2D Gaussians arranged in a square grid, similar to \citet{aithal2024understanding}. Since the true distribution $p(\vx_0)$ is known in this setup, we define hallucinations as samples $\vx$ for which $p(\vx)< p(\vx_{4\sigma})$, where $\vx_{4\sigma}$ is a threshold of 4 standard deviations away from the nearest Gaussian component.
We cluster the generated dataset into 25 clusters and use the cluster assignments as labels for DG.

For the 2D Mixture of Gaussians dataset, as both \citet{aithal2024understanding} and we observe, the number of training data and iterations greatly affect the number of hallucinations, so we train the DDPM \cite{ho2020denoising} for 20k, 50k, and 100k iterations and generate 100k samples with DDPM for evaluation (Table~\ref{table:2dg}). For variance filtering, we evaluate the performance when discarding 1\%, 2.5\%, and 5\% of the generated samples.  We observe that variance filtering requires discarding a very large percentage of samples to match Dynamic Guidance, while our method outperforms it in most cases.

\begin{table}[h]
    \centering
    \small
\caption{2D Mixture of Gaussians dataset.}
\label{table:2dg}
\begin{tabular}{lccc}
\toprule
\multicolumn{1}{c}{\multirow{2}{*}{Method}} & \multicolumn{3}{c}{\# Training iterations}                                     \\ \cmidrule{2-4} 
\multicolumn{1}{c}{}                        & \multicolumn{1}{c}{20k} & \multicolumn{1}{c}{50k} & \multicolumn{1}{c}{100k} \\ \midrule
Variance Filtering (1\%)                      &        6.1\%             & 12.71\%                  &        17.8\%           \\
Variance Filtering (2.5\%)                    &    15.4\%             & 34.5\%                  &   45.7\%                   \\
Variance Filtering (5\%)                      &       34.5\%       & 60.5\%                  &      71.3\%          \\
\textbf{Dynamic Guidance}          &     \textbf{69.5\%}                & \textbf{76.1\%}         &   \textbf{72.1\%}  \\
\bottomrule
\end{tabular}
\end{table}

\subsection{Single Shapes - Mixed Shapes}
For both Shape datasets, we generate 50k images to train the diffusion and classifiers, and sample 10k images using 50-step DDIM for evaluation. Hallucinations are quantified by automatically detecting shapes in the generated images using OpenCV. To evaluate variance filtering, we test thresholds that discard 1\%, 5\%, and 10\% of generated samples. Even at the highest threshold (10\%), variance filtering fails to reliably identify and remove hallucinations. In contrast, Dynamic Guidance consistently mitigates more than 50\% of hallucinations across a broad range of guidance scales, and achieves over a 70\% reduction at the optimal scale (Tables~\ref{table:single_shapes},\ref{table:mixed_shapes}).

\subsection{Hands}
For the hands dataset \citep{afifi201911kHands}, we train both the diffusion models and the classifiers on the 11k images provided. For evaluation, we sample 100 images with each method using 25-step DDIM and manually label each image as hallucinated or not. We perform this experiment 10 times and manually label all 3000 generated images (1000 unguided, 1000 for DG, and 1000 for CG). We report the mean and standard deviation for the experiments. To evaluate variance filtering, we test thresholds that discard 1\%, 5\%, and up to 10\% of generated samples. Even at the highest threshold (10\%), variance filtering fails to reliably identify and remove hallucinations.  DG vastly outperforms both CG and variance filtering, mitigating more than 45\% of hallucinations on average (Table~\ref{table:hands}).

\subsection{ImageNet}

We build on the guided-diffusion codebase\footnote{\url{https://github.com/openai/guided-diffusion}} and implement Dynamic Guidance following Algorithm~\ref{alg:ddim}. We use the pretrained 256x256 unconditional and conditional ADM models and the 256x256 noisy image classifier. For the experiment in Section~\ref{sec:dino} we train the noisy classifier using the training script provided in the guided-diffusion codebase and use the hyperparameters described in \cite{dhariwal2021cg}.

\subsection{Text to Image}
\label{sec:impl_t2i}
We use Stable Diffusion 2.1 with a DDIM scheduler for deterministic sampling. Generation uses 20 denoising steps, guidance scale 7.5, and 512$\times$512 resolution. For all our experiments we use a default guidance window of [850, 300] on a 1000-step schedule. Within this window, we query the VLM every 5 denoising steps.

\paragraph{Classifier VLM Configuration}

We use Qwen2-VL-7B in bfloat16 precision. The VLM receives a multiple-choice prompt with branch options plus a ``STAY'' option for uncertain cases:

\begin{tcolorbox}[colback=gray!5,colframe=gray!75,title=VLM Branch Selection Prompt]
\small
\texttt{Look at this partially-generated image. I need you to determine which description best matches what is ALREADY FORMING in the image.}

\vspace{0.5em}
\texttt{The image is ambiguous about: \{ambiguity\_axis\}}

\vspace{0.5em}
\texttt{Options:}\\
\texttt{A. \{branch\_1\}}\\
\texttt{B. \{branch\_2\}}\\
\texttt{...}\\
\texttt{N. STAY - I cannot tell yet which option best matches the image}

\vspace{0.5em}
\texttt{If none of them seem reasonable or there is no way to tell then pick N to STAY. It's okay to pick a letter even if not completely sure.}

\vspace{0.5em}
\texttt{Respond with ONLY the letter (A, B, C, etc.) and nothing else.}
\end{tcolorbox}

The \texttt{\{ambiguity\_axis\}} placeholder is replaced with the specific ambiguity type (e.g., ``body pose'', ``breed'', ``species''). The branch options are populated from the corresponding branch prompts in Tables ~\ref{tab:prompts1} and ~\ref{tab:prompts2}. The STAY option allows the VLM to defer judgment when the intermediate image is too noisy to classify reliably.

\paragraph{Branch Prompt Design}

For each base prompt, we define a set of branch prompts that resolve the specified ambiguity. We identify a primary axis of semantic ambiguity and generate a set of branch prompts that resolve it. The following prompt was used to generate candidate branches:

\begin{tcolorbox}[colback=gray!5,colframe=gray!75,title=Branch Generation Prompt]
\small
\texttt{Given a text-to-image prompt, identify the primary axis of ambiguity (e.g., body pose, breed, species, object type) and generate branch prompts that resolve it.}

\vspace{0.5em}
\texttt{Requirements:}
\begin{itemize}
    \item \texttt{Each branch should be mutually exclusive}
    \item \texttt{Branches should preserve the base structure while resolving the ambiguity}
    \item \texttt{Keep branches simple and within the model's capability}
    \item \texttt{Avoid low-quality or rare interpretations}
\end{itemize}

\vspace{0.5em}
\texttt{Base prompt: "\{base\_prompt\}"}

\vspace{0.5em}
\texttt{Output format:}\\
\texttt{Ambiguity axis: <axis>}\\
\texttt{Branches:}\\
\texttt{- <branch\_1>}\\
\texttt{- <branch\_2>}\\
\texttt{...}
\end{tcolorbox}

\begin{figure}[t]
    \centering
    \includegraphics[width=\linewidth]{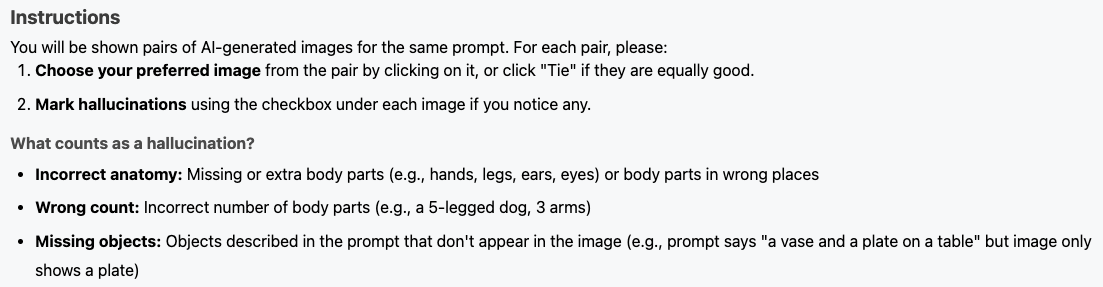}
    \caption{Instructions given to the participants in the user study.}
    \label{fig:instructions}
\end{figure}

Tables~\ref{tab:prompts1} and \ref{tab:prompts2} list all prompts used in the user study. Figure~\ref{fig:instructions} shows the instructions given to the participants.

\begin{table}[h]
\centering
\caption{Base prompts and their corresponding branch prompts used in the user study (Part 1/2).}
\label{tab:prompts1}
\small
\begin{tabular}{p{3cm}p{2cm}p{8cm}}
\toprule
\textbf{Base Prompt} & \textbf{Ambiguity} & \textbf{Branch Prompts} \\
\midrule

a cat in a living room & body pose &
\begin{minipage}[t]{8cm}
\begin{itemize}
    \item a cat sitting upright in a living room
    \item a cat lying down curled up in a living room
    \item a cat standing on all fours in a living room
    \item a cat stretching with arched back in a living room
    \item a cat walking across a living room
    \item a cat grooming itself in a living room
    \item a cat jumping in a living room
    \item a cat scratching furniture in a living room
\end{itemize}
\end{minipage} \\
\midrule

a dog playing fetch & body pose &
\begin{minipage}[t]{8cm}
\begin{itemize}
    \item a dog running with a ball in its mouth
    \item a dog jumping to catch a ball
    \item a dog waiting for a ball to be thrown
    \item a dog chasing a ball
    \item a dog returning with a ball
    \item a dog dropping a ball at feet
\end{itemize}
\end{minipage} \\
\midrule

a cat sleeping & breed &
\begin{minipage}[t]{8cm}
\begin{itemize}
    \item a persian cat sleeping
    \item a siamese cat sleeping
    \item a maine coon cat sleeping
    \item a tabby cat sleeping
    \item an orange cat sleeping
    \item a black cat sleeping
    \item a white cat sleeping
    \item a gray cat sleeping
    \item a calico cat sleeping
\end{itemize}
\end{minipage} \\
\midrule

a dog on a couch & breed &
\begin{minipage}[t]{8cm}
\begin{itemize}
    \item a poodle on a couch
    \item a husky on a couch
    \item a dachshund on a couch
    \item a bulldog on a couch
    \item a corgi on a couch
    \item a chihuahua on a couch
    \item a golden retriever on a couch
    \item a shiba inu on a couch
    \item a pit bull on a couch
\end{itemize}
\end{minipage} \\
\midrule

a bird on a branch & species &
\begin{minipage}[t]{8cm}
\begin{itemize}
    \item a robin on a branch
    \item a blue jay on a branch
    \item a cardinal on a branch
    \item a sparrow on a branch
    \item a crow on a branch
    \item a parrot on a branch
    \item an owl on a branch
    \item a woodpecker on a branch
\end{itemize}
\end{minipage} \\

\bottomrule
\end{tabular}
\end{table}

\begin{table}[h]
\centering
\caption{Base prompts and their corresponding branch prompts used in the user study (Part 2/2).}
\label{tab:prompts2}
\small
\begin{tabular}{p{3cm}p{2cm}p{8cm}}
\toprule
\textbf{Base Prompt} & \textbf{Ambiguity} & \textbf{Branch Prompts} \\
\midrule
a boat on water & vessel type &
\begin{minipage}[t]{8cm}
\begin{itemize}
    \item a sailboat on water
    \item a speedboat on water
    \item a rowboat on water
    \item a fishing boat on water
    \item a yacht on water
    \item a canoe on water
    \item a kayak on water
\end{itemize}
\end{minipage} \\
\midrule

a cat on a windowsill & body pose &
\begin{minipage}[t]{8cm}
\begin{itemize}
    \item a cat sitting upright on a windowsill
    \item a cat lying stretched out on a windowsill
    \item a cat curled up on a windowsill
    \item a cat looking out the window on a windowsill
    \item a cat sleeping on a windowsill
\end{itemize}
\end{minipage} \\
\midrule

a dog by a fireplace & body pose &
\begin{minipage}[t]{8cm}
\begin{itemize}
    \item a dog sleeping by a fireplace
    \item a dog sitting by a fireplace
    \item a dog lying on its side by a fireplace
    \item a dog with head on paws by a fireplace
    \item a dog stretching by a fireplace
\end{itemize}
\end{minipage} \\
\midrule

a frog & pose &
\begin{minipage}[t]{8cm}
\begin{itemize}
    \item a frog sitting on a lily pad
    \item a frog jumping
    \item a frog in water
    \item a frog on a rock
\end{itemize}
\end{minipage} \\
\midrule

a rabbit & pose &
\begin{minipage}[t]{8cm}
\begin{itemize}
    \item a rabbit sitting upright
    \item a rabbit hopping
    \item a rabbit lying down
    \item a rabbit eating
    \item a rabbit in grass
\end{itemize}
\end{minipage} \\

\bottomrule
\end{tabular}
\end{table}

\subsection{Compute details}
\label{sec:compute_details}
For all experiments, we use a cluster of NVIDIA RTX 6000 Ada GPUs. For the discrete conditioning models (Shapes, ImageNet), Dynamic Guidance requires training a noisy image classifier. This incurs some computational overhead during training, which, however, is significantly lower than that of training the Diffusion Model itself. We trained the classifier on pseudo-classes for the experiment of Section~\ref{sec:dino} on 4 NVIDIA RTX 6000 Ada GPUs for 50 hours. For the rest of the experiments we used pre-trained publicly availanble checkpoints. For a more extensive analysis of the compute required to train ADM models and classifiers, we refer to Appendix A of \cite{dhariwal2021cg}.

During generation, when using discrete condition models, DG requires a forward and a backward pass through the classifier, exactly like CG, for every step that it is active. This means that it requires some computational overhead compared to the unguided model. Crucially, compared to CG, DG can be applied to a subset of generation steps only, so it incurs a lower computational cost than CG. To compare the three settings, we sample 160 256x256 images with the unguided model, DG and CG in a single NVIDIA RTX 6000 Ada and compare the generation time in Table~\ref{tab:generation_time}.

\begin{table}[H]
\caption{Comparison of generation times for different guidance methods.}
\centering
\begin{tabular}{lc}
\toprule
\textbf{Method} & \textbf{Generation time (s)} \\
\midrule
ADM & 351 \\
Classifier Guidance (applied to all timesteps) & 417 \\
Dynamic Guidance (applied to timesteps 800--400) & 380 \\
\bottomrule
\end{tabular}
\label{tab:generation_time}
\end{table}

\clearpage
\section{Additional Comparisons / Metrics}
\label{fid}

\begin{figure}[ht]
    \centering
    \includegraphics[width=\linewidth]{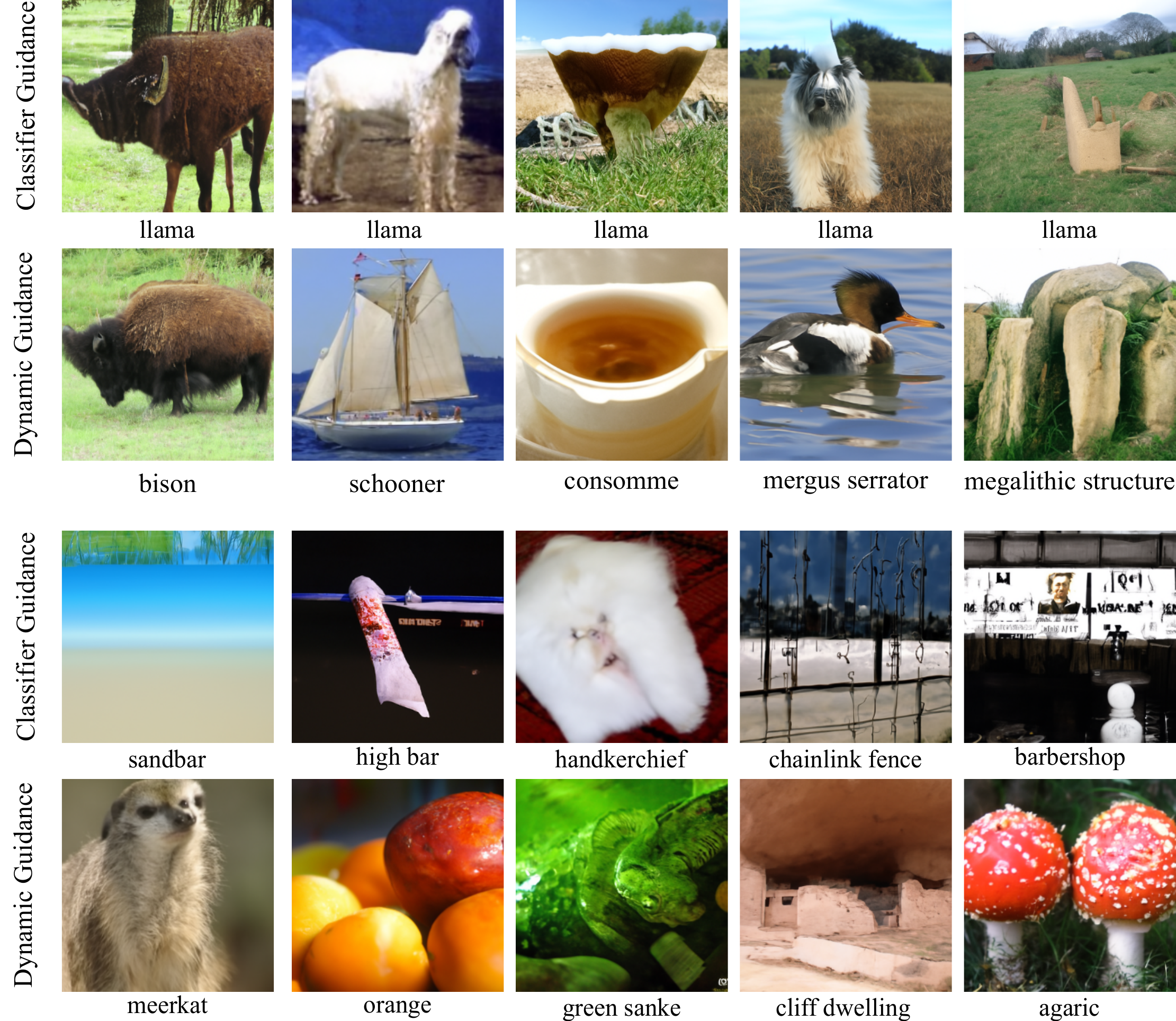}
    \caption{Images generated with Classifier and Dynamic Guidance using the same initial noises. \textbf{(Top)} Given a specific label (``llama") that is misaligned with the initial noise, the model can generate low-quality samples that visually resemble hallucinations. \textbf{(Bottom)} Even for randomly selected labels, when those are fixed, misalignment can produce generations that would be considered hallucinations.}
    \label{fig:imagenet}
\end{figure}

\begin{table}[H]
\centering
\small
\caption{Performance of Classifier (CG) and Dynamic Guidance (DG) on Hallucination-related metrics on ImageNet-1k generation.}
\label{table:imagenet}
\begin{tabular}{lccccc}
\toprule
&  \multicolumn{3}{c}{Inception} & \multicolumn{2}{c}{CLIP} \\
\cmidrule(lr){2-4} \cmidrule(lr){5-6}
Method & IS$\uparrow$ & Prec$\uparrow$ & Rec$\uparrow$ & Prec$\uparrow$ & Rec$\uparrow$ \\
\midrule
Uncond. ADM          & 34.24 & 0.53 & 0.61 & 0.60 & 0.26 \\
\midrule
+ CG $\lambda=1$       & 28.52 & 0.47 & 0.62 & 0.57 & 0.25 \\
\textbf{+ DG $\lambda=1$} & 49.63 & 0.62 & 0.59 & 0.66 & \textbf{0.27} \\
\midrule
+ CG $\lambda=2$& 38.39 & 0.55 & \textbf{0.64} & 0.63 & \textbf{0.27} \\
\textbf{+ DG $\lambda=2$} & 59.31 & 0.70 & 0.56 & 0.70 & 0.26 \\
\midrule
+ CG $\lambda=10$    & 83.19 & 0.69 & 0.55 & 0.74 & \textbf{0.27} \\
\textbf{+ DG $\lambda=10$} & \textbf{88.49} & \textbf{0.77} & 0.52 & \textbf{0.77} & 0.26 \\
\midrule
+ CG $\lambda=20$    & 75.84 & 0.62 & 0.55 & 0.70 & 0.23 \\
\textbf{+ DG $\lambda=20$} & 80.99 & 0.70 & 0.52 & 0.75 & 0.23 \\
\bottomrule
\end{tabular}
\end{table}

We also compute Inception and CLIP FID and observe that in most settings Dynamic Guidance outperforms Classifier Guidance. For high guidance scales ($\lambda=10$), Dynamic Guidance tends to generate classes that are better matched with a wider range of initial noises, and so it ends up creating a distribution of generated samples that is non-uniform with regard to the ImageNet classes. This greatly affects Inception FID since the Inception model is heavily biased towards balanced ImageNet generation (trained on ImageNet-1k), whereas performance on CLIP FID is not affected. To fairly compare Dynamic Guidance to Classifier Guidance that strictly enforces a balanced generated distribution for evaluation, we generate a larger amount of samples and perform stratified sampling on the generated set to approximate a balanced distribution. We report the results in Table~\ref{fid_table} in the row titled \emph{+ DG $\lambda=10$ (balanced)}.

\begin{table}[H]
\centering
\begin{minipage}[t]{0.5\linewidth}
\centering
\small
\caption{Inception and CLIP FID on ImageNet.}
\begin{tabular}{lcc}
\toprule
 & \multicolumn{2}{c}{FID} \\
\cmidrule(lr){2-3}
Method & Inception$\downarrow$ & CLIP$\downarrow$ \\
\midrule
Uncond. ADM          & 37.48 & 42.93 \\
\midrule
+ CG $\lambda=1$       & 43.72 & 48.02 \\
+ DG $\lambda=1$ & 27.46 & 38.21 \\
\midrule
+ CG $\lambda=2$       & 32.96 & 41.15 \\
+ DG $\lambda=2$ & 25.36 & 36.66 \\
\midrule
+ CG $\lambda=10$      & 17.05 & 30.01 \\
+ DG $\lambda=10$ & 25.57 & 32.68 \\
\textbf{+ DG $\mathbf{\lambda=10}$ (balanced)} & \textbf{15.52} & \textbf{27.93} \\
\midrule
+ CG $\lambda=20$      & 21.45 & 40.50 \\
+ DG $\lambda=20$ & 25.58 & 37.07 \\
\bottomrule
\end{tabular}
\label{fid_table}
\end{minipage}
\hspace{1cm}
\begin{minipage}[t]{0.4\linewidth}
    \centering
    \small
    \caption{Density and Coverage on ImageNet-1k generations, based on CLIP.}
    \label{table:imagenet_density_coverage}
    \begin{tabular}{lcc}
    \toprule
    \textbf{Method} & \textbf{Density} $\uparrow$ & \textbf{Coverage} $\uparrow$ \\
    \midrule
    Uncond. & 0.7003 & 0.6662 \\
    \midrule
    +CG ($\lambda=1$)  & 0.5366 & 0.4473 \\
    +DG ($\lambda=1$)  & 0.7023 & 0.5921 \\
    \midrule
    +CG ($\lambda=10$) & 0.9058 & 0.7516 \\
    +DG ($\lambda=10$) & 0.9724 & 0.6761 \\
    \midrule
    +CG ($\lambda=20$) & 0.7999 & 0.6943 \\
    +DG ($\lambda=20$) & 0.9285 & 0.6385 \\
    \bottomrule
    \end{tabular}
\end{minipage}
\end{table}

We also report additional metrics aiming to better illustrate the trade-off between reducing hallucinations and loss of diversity in generation. We compute diversity and coverage, as described by \citet{naeem2020reliable} using CLIP, and report the results in Table~\ref{table:imagenet_density_coverage}. We see that DG improves the density of the generations from 0.70 to 0.97 while not sacrificing coverage (0.66 vs 0.67).

\subsection{Additional Analysis of Dynamic Guidance}
\begin{figure}[h]
    \centering
    \includegraphics[width=\linewidth]{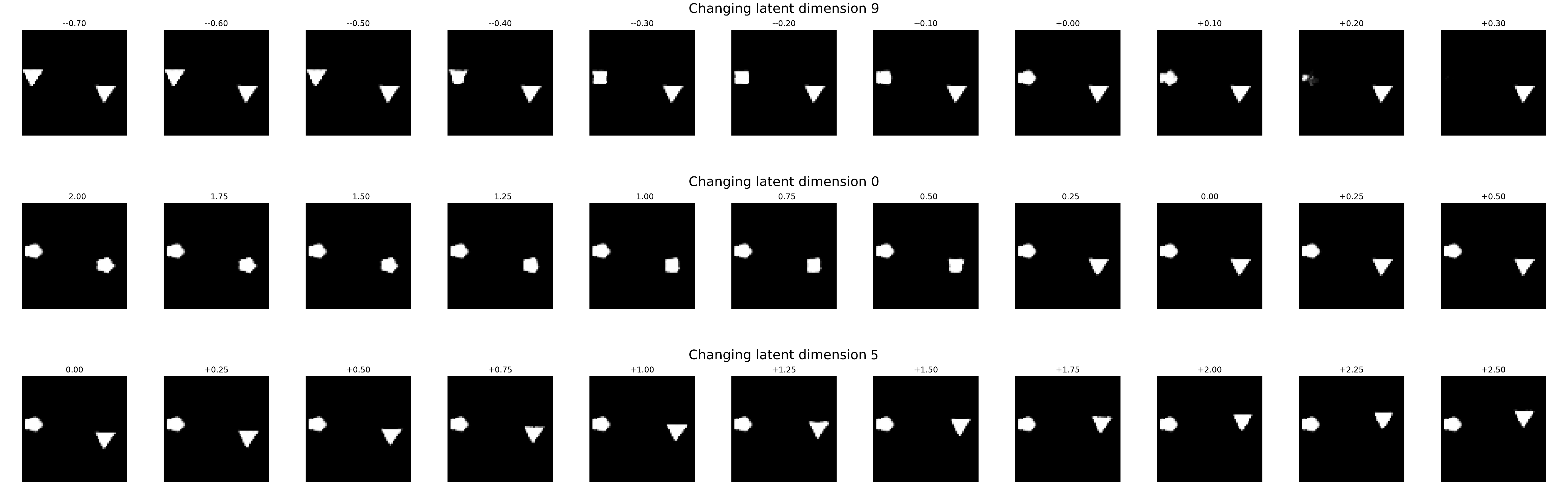}
    \includegraphics[width=1.0\linewidth]{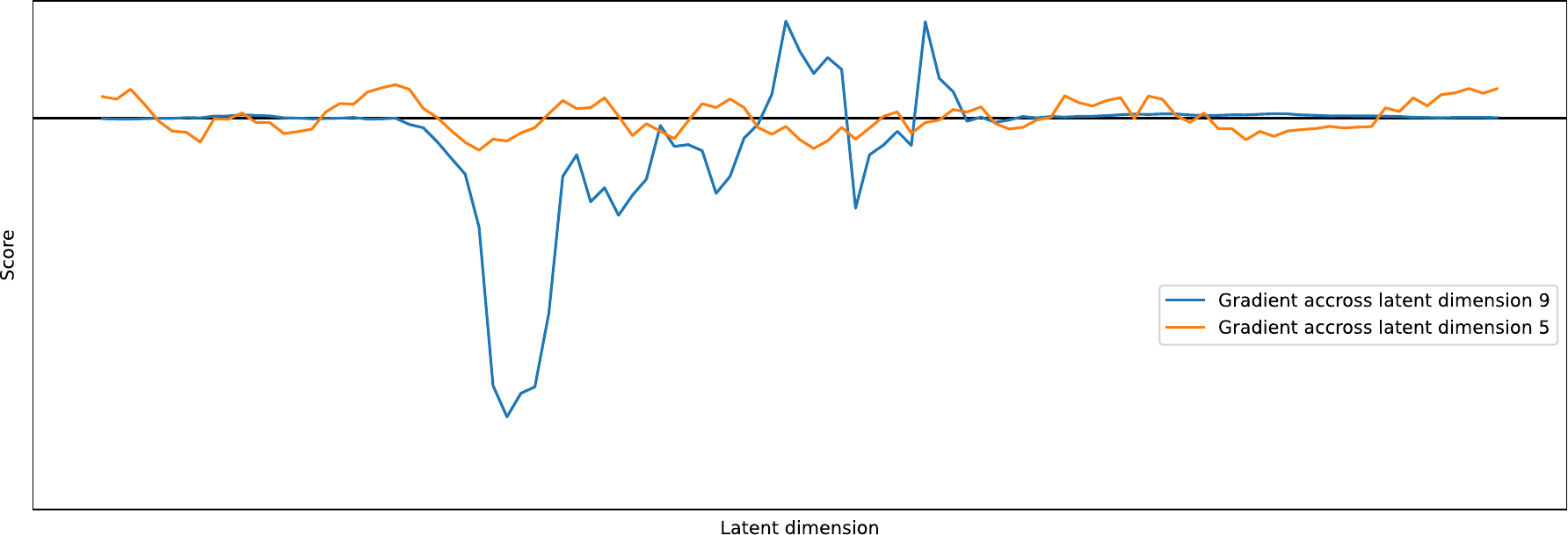}
    \caption{(a) Change in different latent dimensions learned by the $\beta$-{VAE} alters the image in distinct ways; latent dimension 9 controls the appearance of the shape on the left, latent dimension 0 controls the appearance of the shape on the right, and latent dimension 5 controls the position of the shape on the right. (b) Dynamic Guidance isolates latent dimensions corresponding to hallucinations (dimension 9), while not affecting unrelated ones (dimension 5), as the gradients along hallucination-relevant directions are strong and informative, whereas gradients along irrelevant directions are noisy and close to zero. This is more pronounced when classes are selected so that interpolation between them directly aligns with hallucinations.}
    \label{fig:latent_traversal}
\end{figure}

\paragraph{Score Sharpening and Guidance Gradients.}
We identify that our trained $\beta$-{VAE} has learned latent dimensions that change the image in distinct, independent ways. We provide visual examples in Figure~\ref{fig:latent_traversal}(a): latent dimensions 0 and 9 control the appearence of shapes placed in different positions in the image, while other dimensions, like 5, control the position of shapes. We show that Dynamic Guidance isolates latent dimensions corresponding to hallucinations (dimension 9), while not affecting unrelated ones (dimension 5), as the gradients along hallucination-relevant directions are strong and informative, whereas gradients along irrelevant directions are noisy and close to zero (Figure~\ref{fig:latent_traversal}(b)). This is more pronounced when classes are selected so that interpolation between them directly aligns with hallucinations, like in the setting of the Single Shapes dataset.

\subsection{Hallucination Mitigation during Sampling}

\begin{figure}[t]
    \centering
    \includegraphics[width=\linewidth]{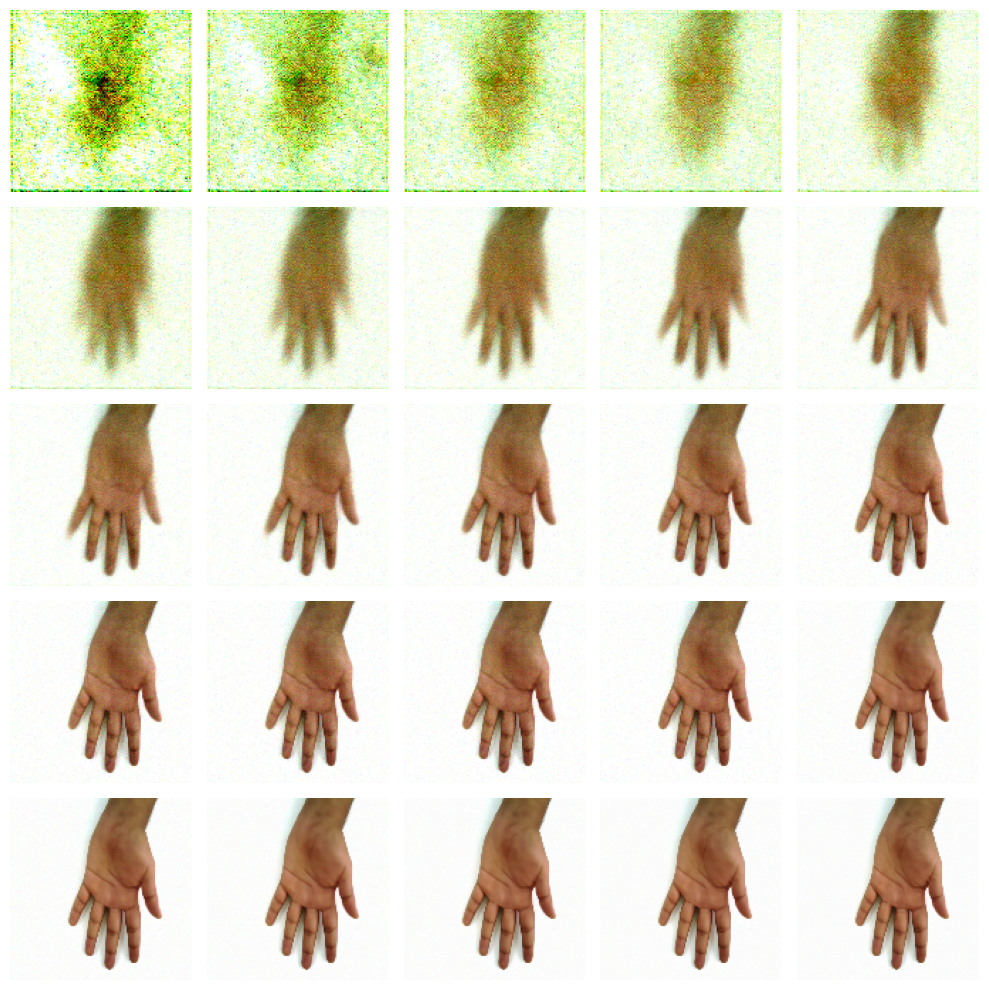}
    \caption{$\hat{\vx}_0$ during sampling with Dynamic Guidance using initial noise $\vx_T$ that would result in a hallucination. We see that Dynamic Guidance guides the model to generate the thumb that would otherwise be missing, resulting in a sample with correct anatomy.}
    \label{fig:steps_hands}
\end{figure}

\begin{figure}[t]
    \centering
    \includegraphics[width=0.45\linewidth]{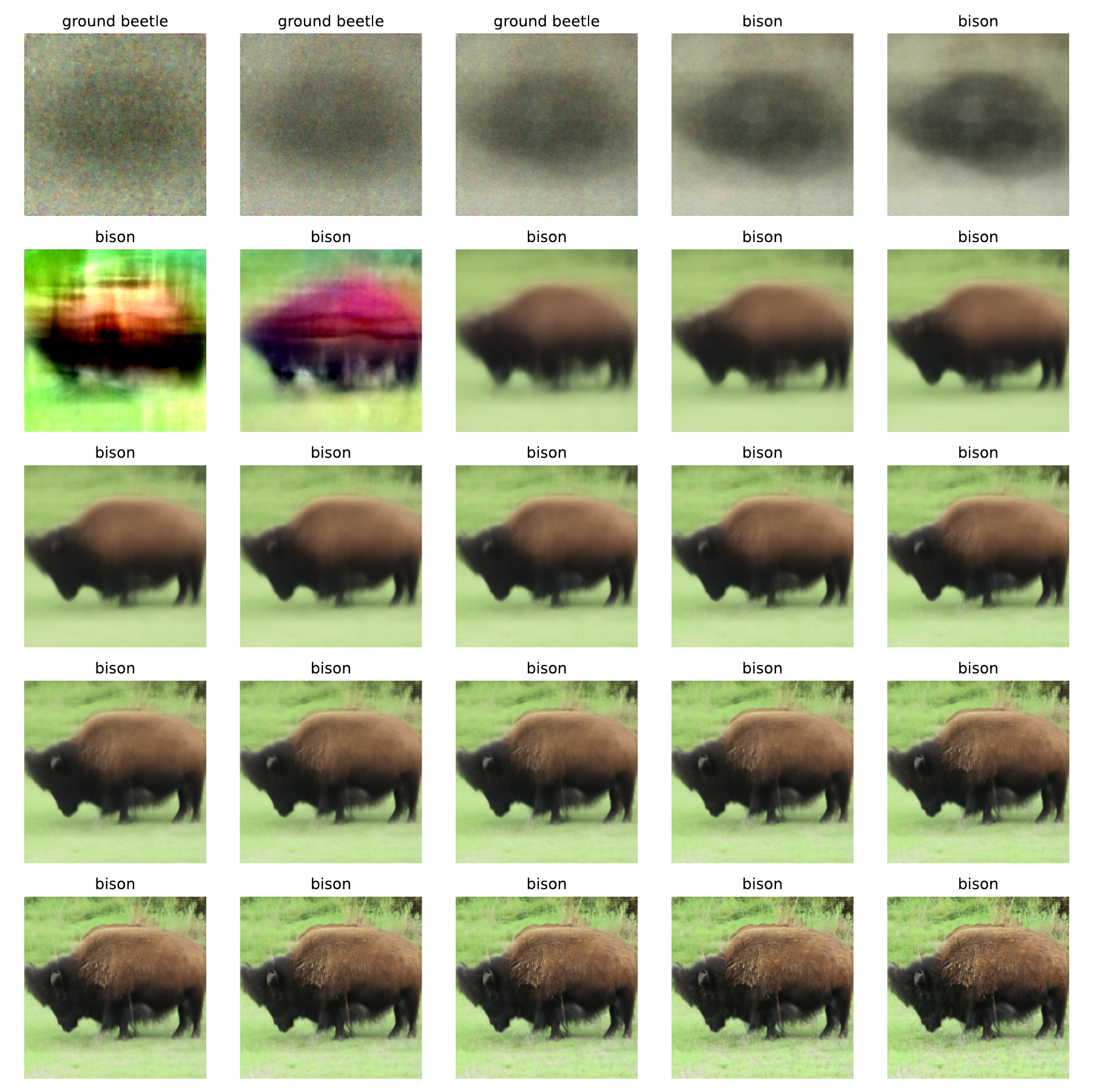}
    \includegraphics[width=0.45\linewidth]{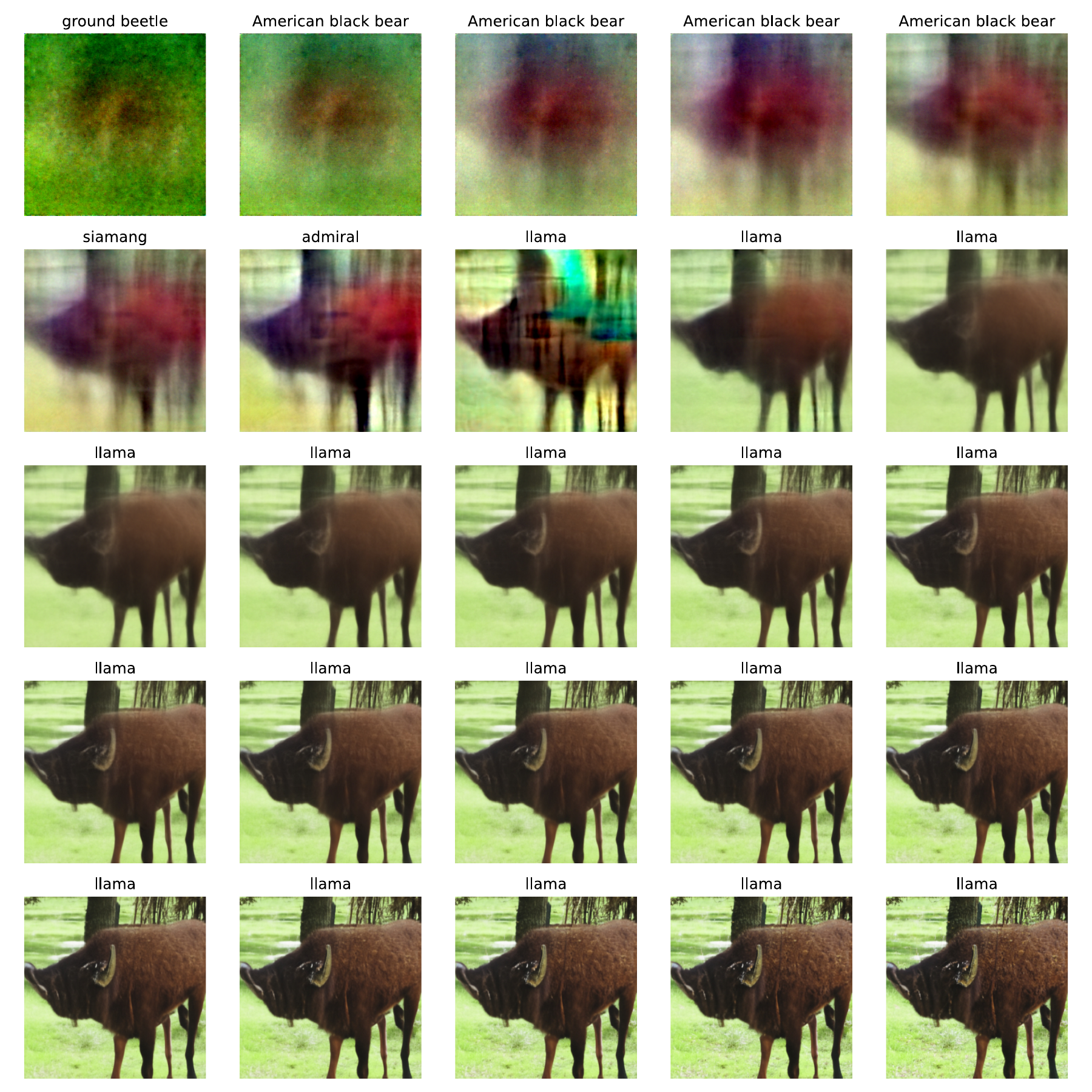}
    \caption{(a) DG steps: When using dynamic guidance, the generated image has the label ``bison". The initial noise adds a blob of black pixels in the middle on a green background, for which the local mode is a bison, as predicted by the classifier. By applying the gradients from dynamic guidance, we end up at a realistic-looking bison image, where both the model and classifier agree.
    (b) CG steps: For the same initial noise, we choose a class that is unlikely to contain a large blob of black pixels in the middle (e.g., ``llama"). We see that the classifier guidance gradient updates attempt to correct the image by changing the shape of the generated object and adding thin legs.}
    \label{fig:bisonvllama}
\end{figure}

To understand how hallucinations form during sampling and how Dynamic Guidance changes the sampling process to mitigate them, we visualize the sampling process for a sample that would otherwise be a hallucination and observe how Dynamic Guidance corrects it (Figure~\ref{fig:steps_hands}). We also show an example where classifier guidance fails to fix a hallucinated sample and how it is fixed using Dynamic Guidance (Figure~\ref{fig:bisonvllama}).

\clearpage
\subsection{Importance of Adaptive Selection}
\begin{figure}[H]
    \centering
    \includegraphics[width=\linewidth]{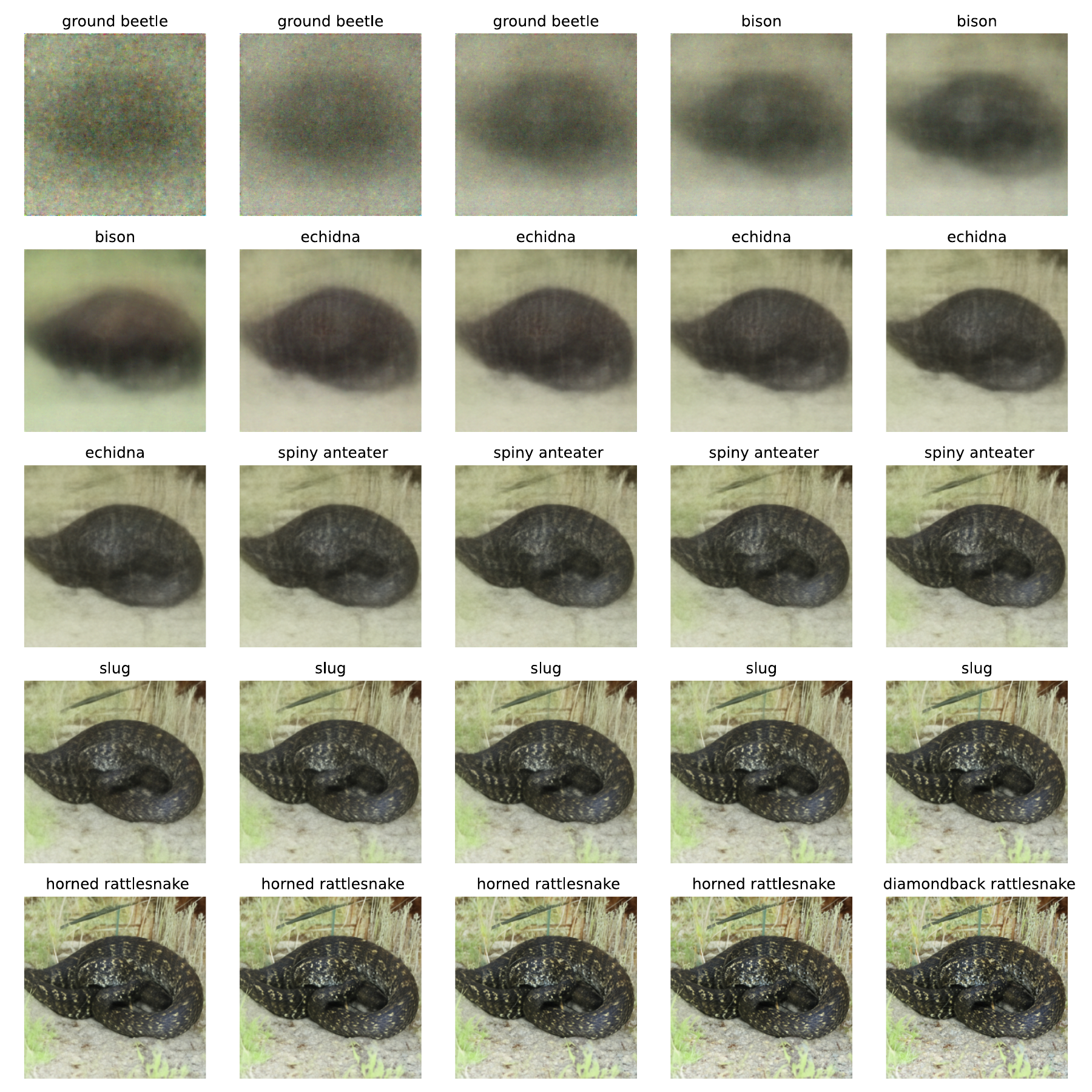}
    \caption{The intermediate image can be close to multiple modes during generation. The selected mode can change multiple times during sampling, and Dynamic Guidance adaptively chooses the closest one.}
    \label{fig:echidna}
\end{figure}

\begin{figure}[H]
    \centering
    \includegraphics[width=0.5\linewidth]{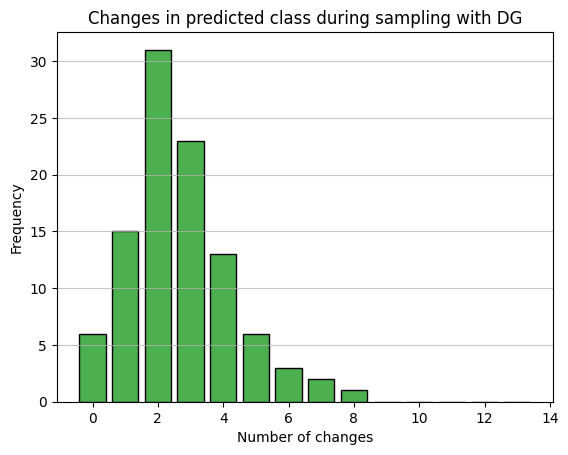}
    \caption{Histogram for the number of selected class changes during generations.}
    \label{fig:changes}
\end{figure}

To highlight the importance of dynamically adjusting the guidance signal we investigate how often the dominant class changes during sampling in Figure~\ref{fig:echidna}. We show that the dominant class can change multiple times during the intermediate steps where we perform the guidance. The class changes more in the early timesteps, where the image has not fully formed yet, and multiple candidate classes could potentially be feasible generations.
To quantify this, we generate 100 images with an unconditional ADM trained on ImageNet and Dynamic Guidance and observe how often the dominant class changes for 25-step generation on average (Figure~\ref{fig:changes}). We observe that for most samples the label changes at least 2 times, which is expected, as for early timesteps the image has not formed fully yet and the sample could feasibly belong to multiple similar classes.

\subsection{Sensitivity to Classifier Quality}
\begin{figure}[H]
    \centering
    \includegraphics[width=\linewidth]{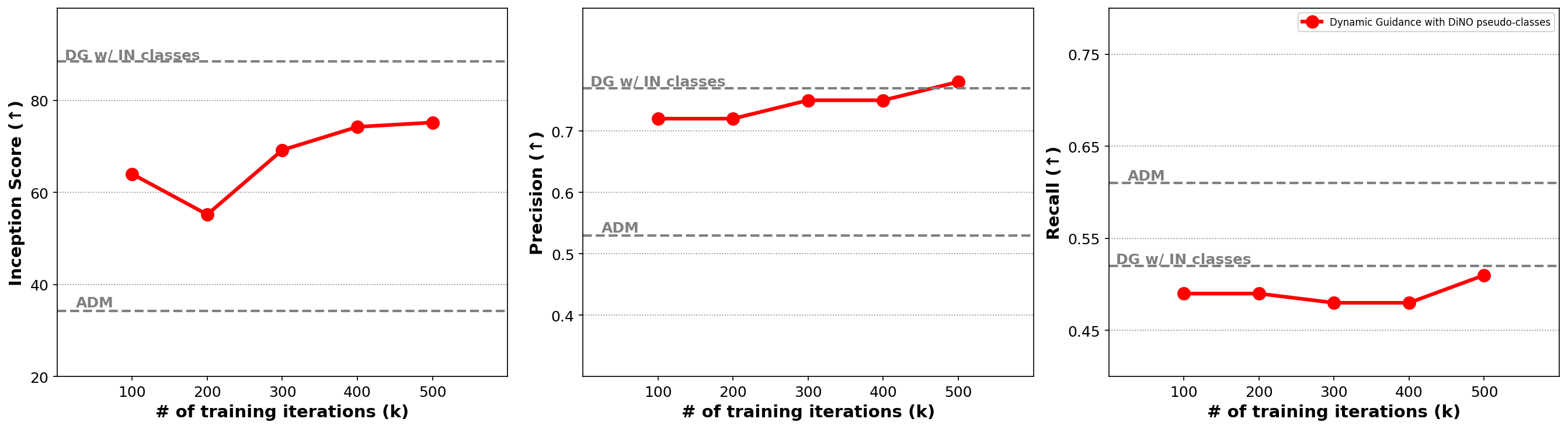}
    \caption{Hallucination related metrics for different training iterations.}
    \label{fig:weak_class}
\end{figure}

To assess the dependence of Dynamic Guidance on specific strong classifiers, we evaluate using weaker classifiers corresponding to earlier training checkpoints. In Figure~\ref{fig:weak_class} we provide hallucination reduction metrics on ImageNet using DG with checkpoints of the DiNO pseudo-class classifier during different stages of its training.
We observe that DG performs well even with a weaker classifier and is effective in mitigating hallucinations (high precision). The gradients provided by the classifier are informative enough to guide the generation away from hallucinations, even when the classifier is slightly worse at differentiating between pseudo-classes or its predictions are less certain.

\subsection{Potential Class Bias in Generation}
\begin{figure}[t]
    \centering
    \includegraphics[width=1.0\linewidth]{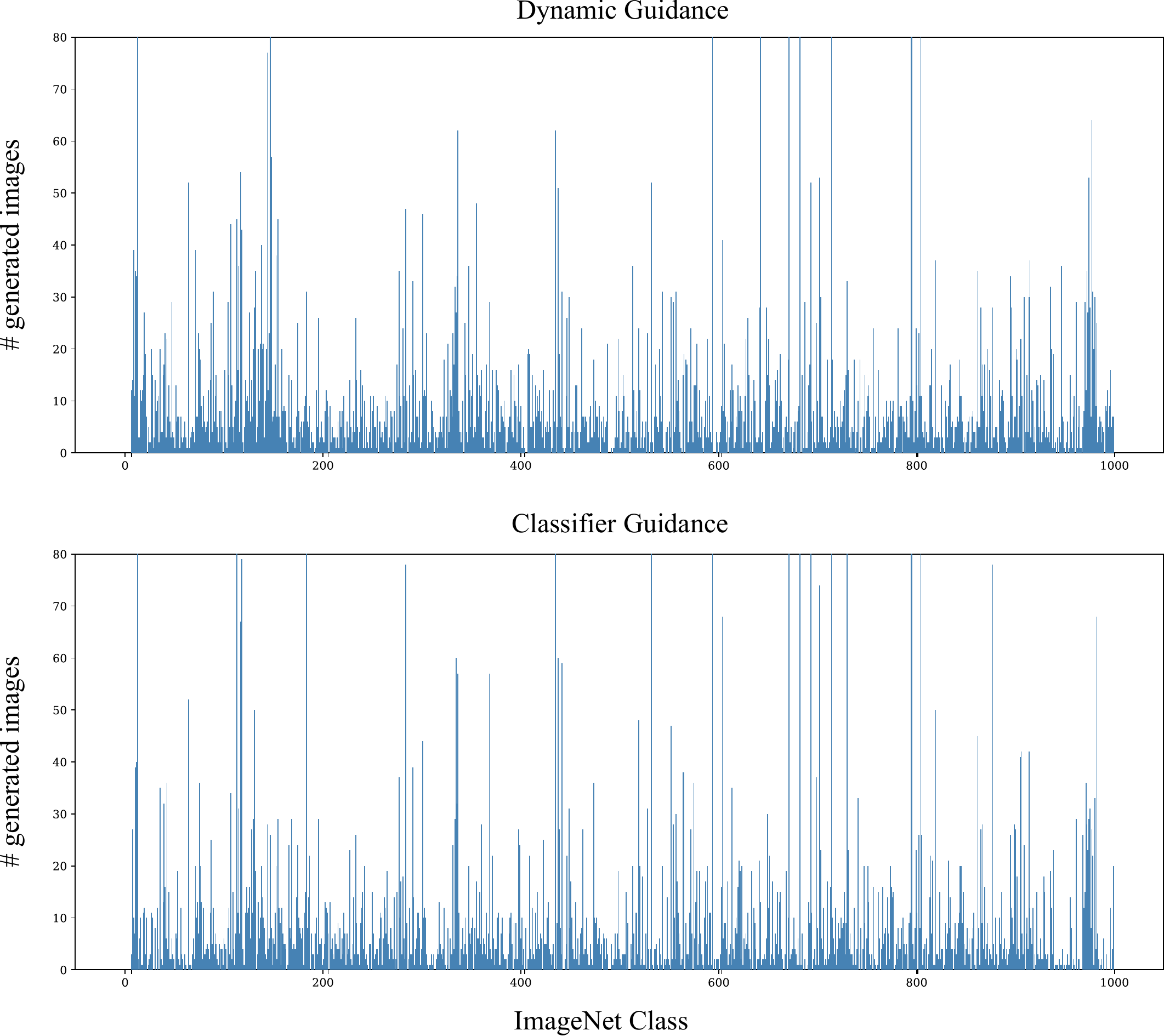}
    \caption{Distribution of final predicted ImageNet classes for samples generated with Classifier and Dynamic Guidance using $\lambda=1$.}
    \label{fig:gen_classes_1}
\end{figure}

\begin{figure}[t]
    \centering
    \includegraphics[width=1.0\linewidth]{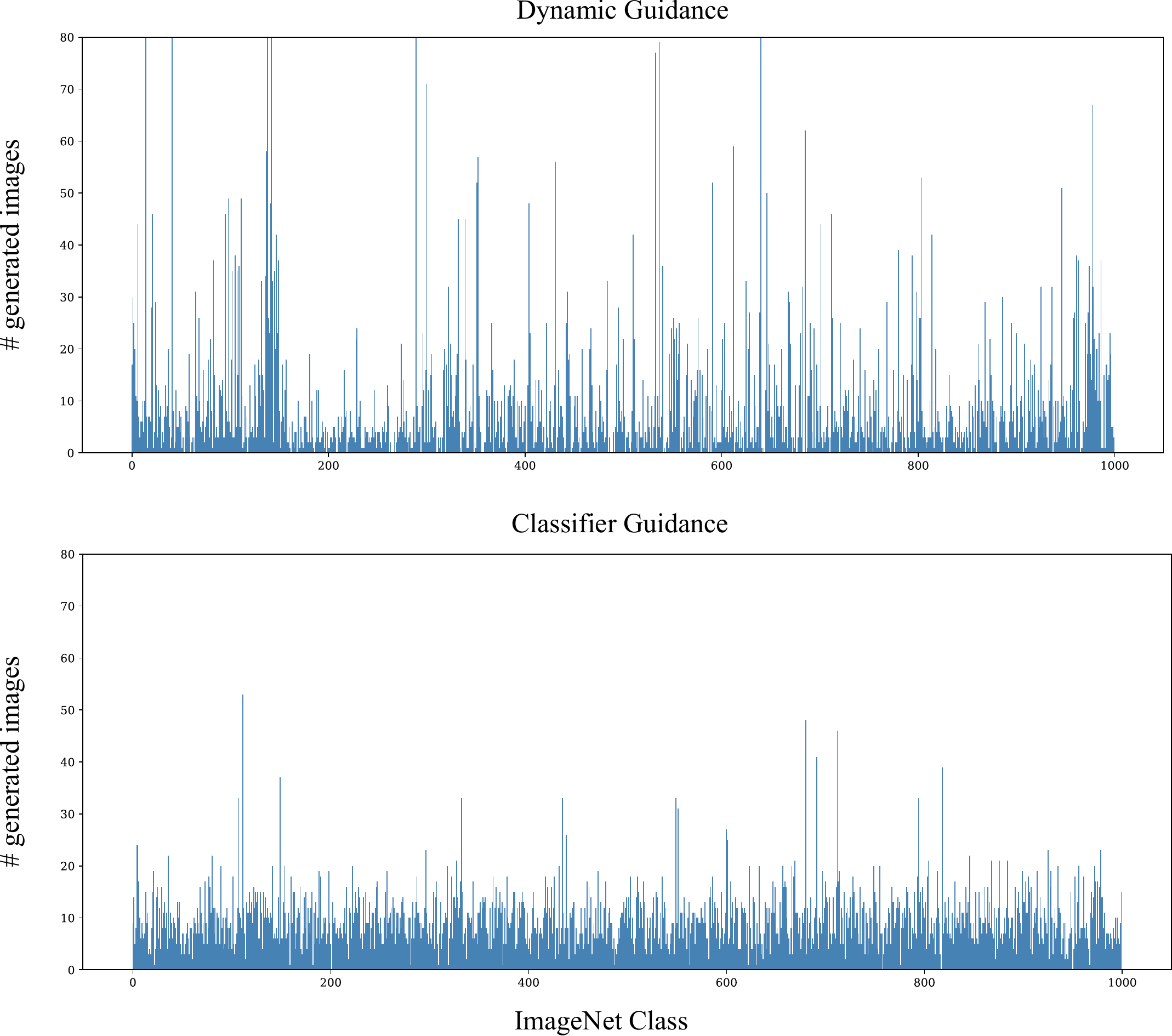}
    \caption{Distribution of final predicted ImageNet classes for samples generated with Classifier and Dynamic Guidance using $\lambda=10$.}
    \label{fig:gen_classes_10}
\end{figure}

To better understand the potential bias introduced by the noisy image classifier, we plot the final classification for the generated images using Classifier Guidance and Dynamic Guidance for different guidance scales (Figures~\ref{fig:gen_classes_1} and \ref{fig:gen_classes_10}).

\subsection{Additional Qualitative Results}
\begin{figure}[t!]
    \centering
    \includegraphics[width=\linewidth]{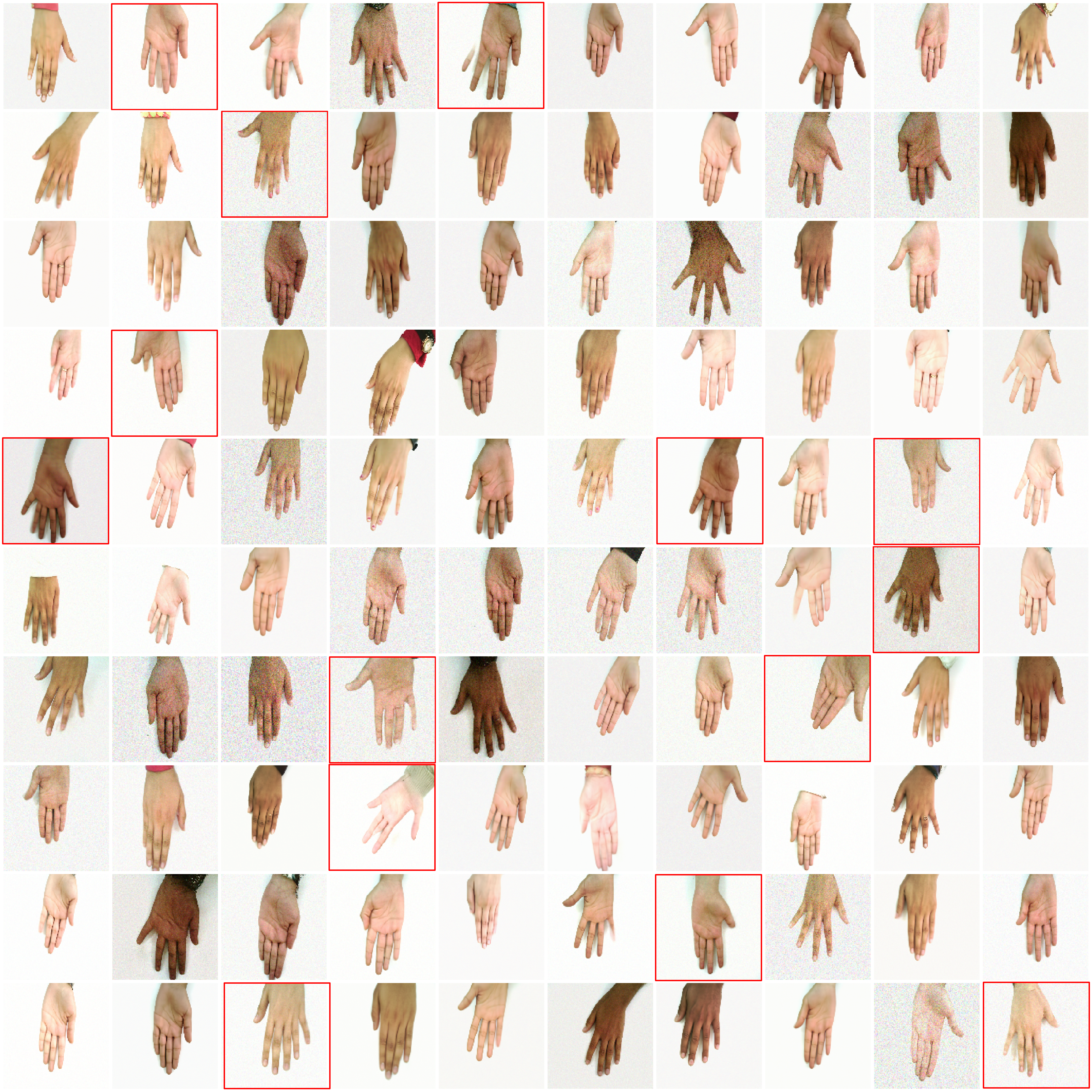}
    \caption{Generated samples from the hands dataset using DDIM. Hallucinations in red.}
    \label{fig:hands_ddim}
\end{figure}

\begin{figure}[t!]
    \centering
    \includegraphics[width=\linewidth]{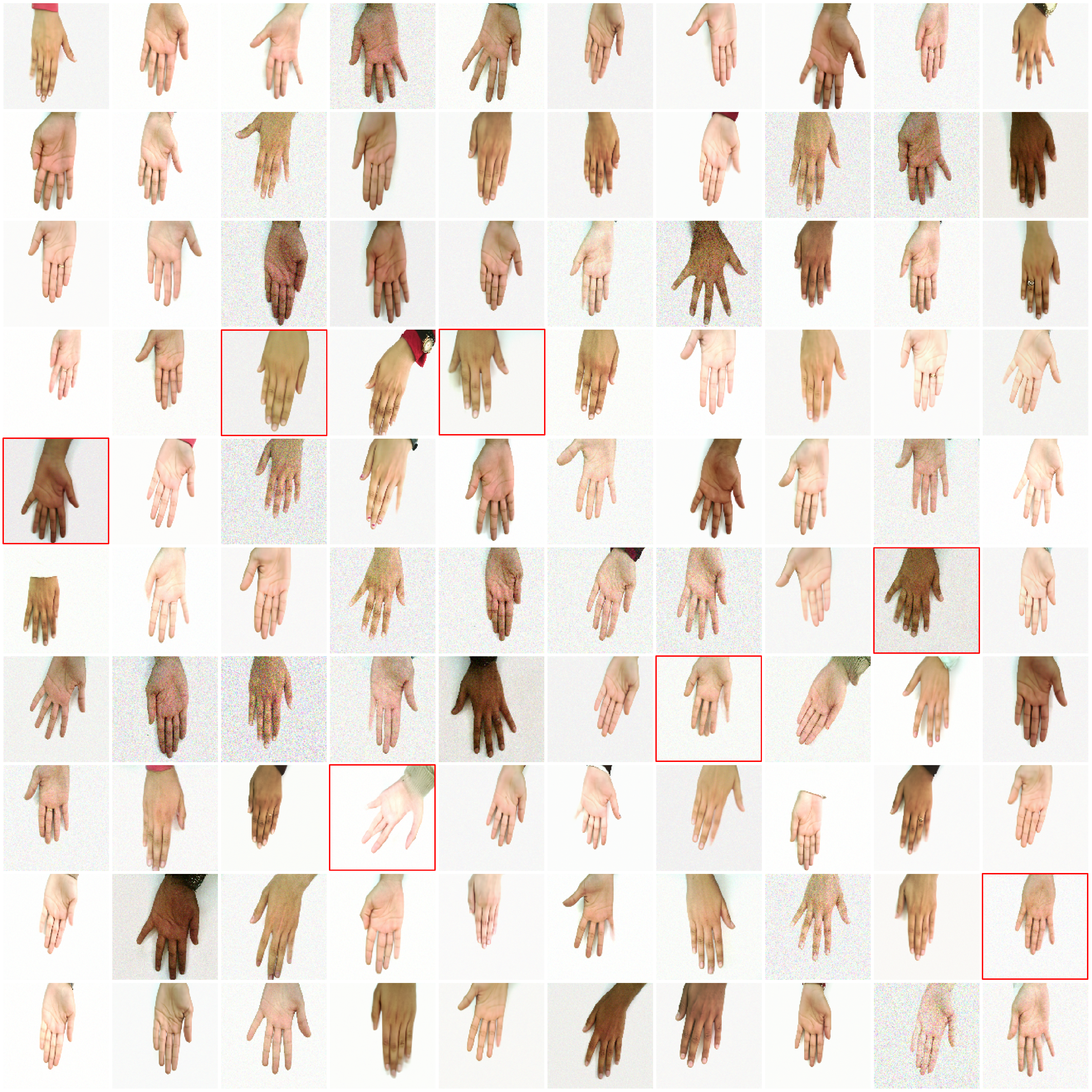}
    \caption{Generated samples from the hands dataset using Dynamic Guidance. Hallucinations in red.}
    \label{fig:hands_dmlcg}
\end{figure}

\begin{figure}[t!]
    \centering
    \includegraphics[width=0.9\linewidth]{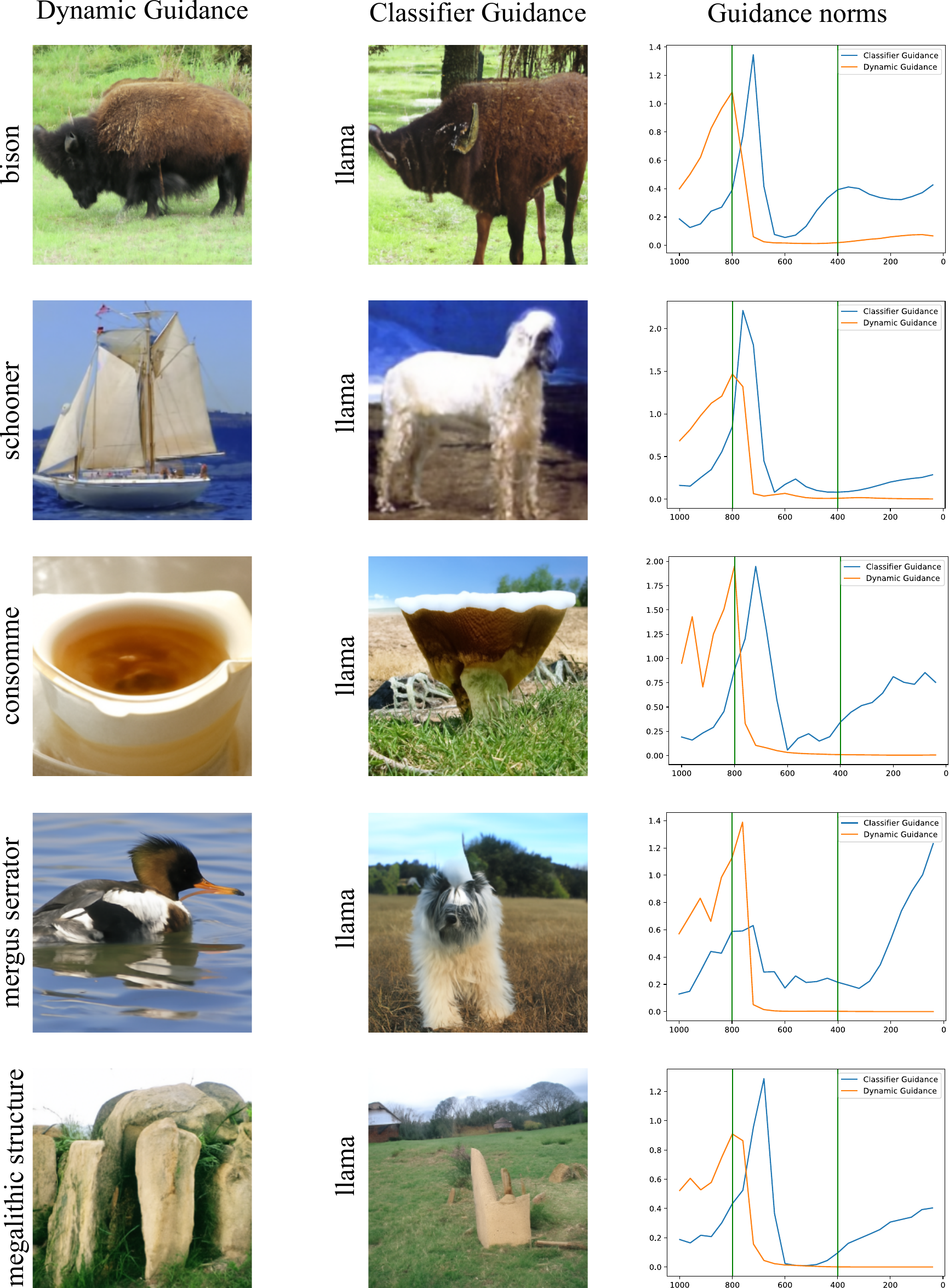}
    \caption{Generated samples from ImageNet using Classifier Guidance with a fixed label and Dynamic Guidance. In classifier-guided samples, the norm of the denoising step gets bigger towards the end of sampling, meaning that the required denoising steps become significantly larger.}
    \label{fig:norm_llama}
\end{figure}

\begin{figure}[t!]
    \centering
    \includegraphics[width=0.9\linewidth]{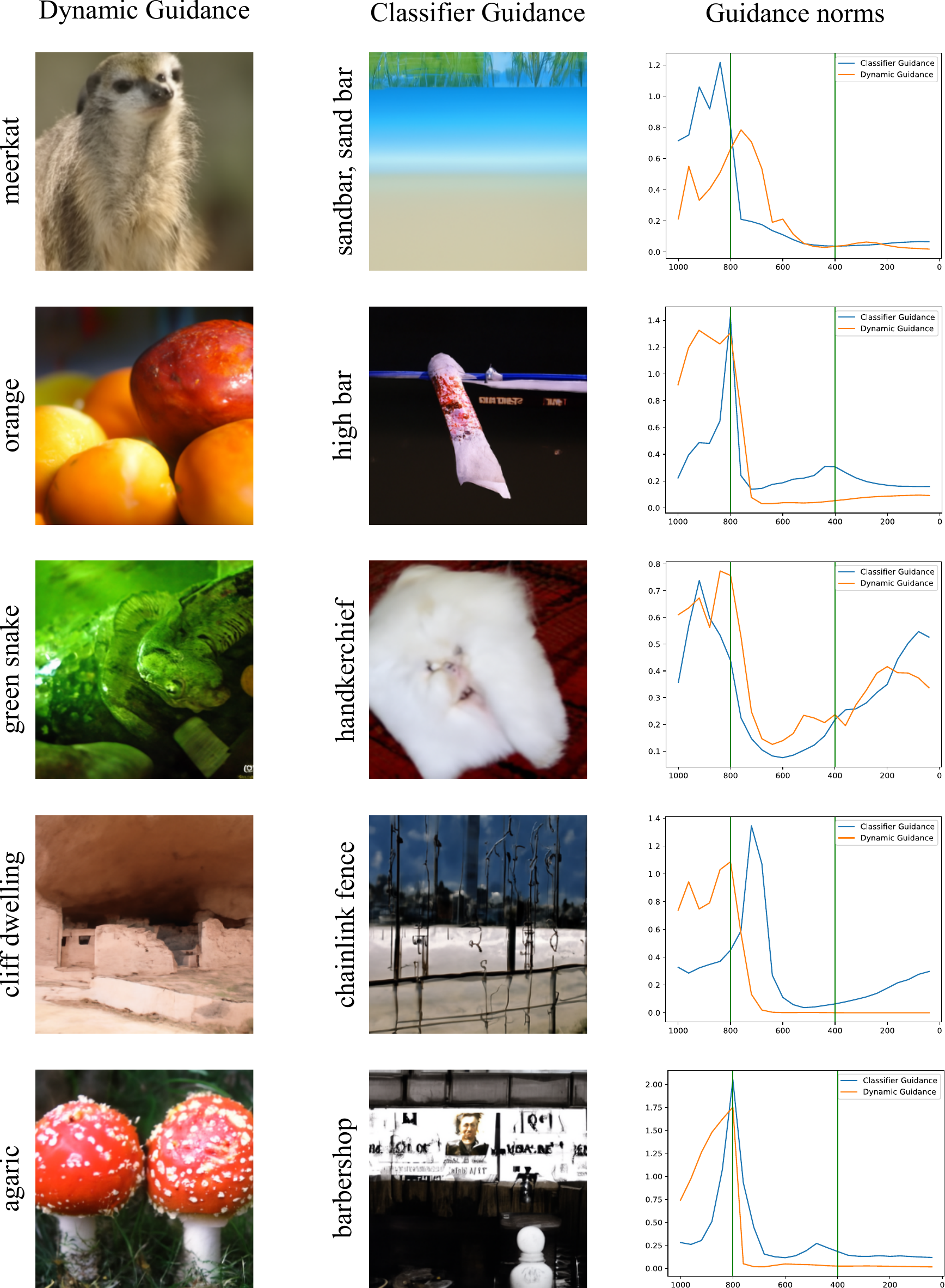}
    \caption{Generated samples from ImageNet using Classifier Guidance with a random label and Dynamic Guidance. In classifier-guided samples, the norm of the denoising step gets bigger towards the end of sampling, meaning that the required denoising steps become significantly larger.}
    \label{fig:norm_random}
\end{figure}

\begin{figure}[t]
    \centering
    \includegraphics[width=.7\linewidth]{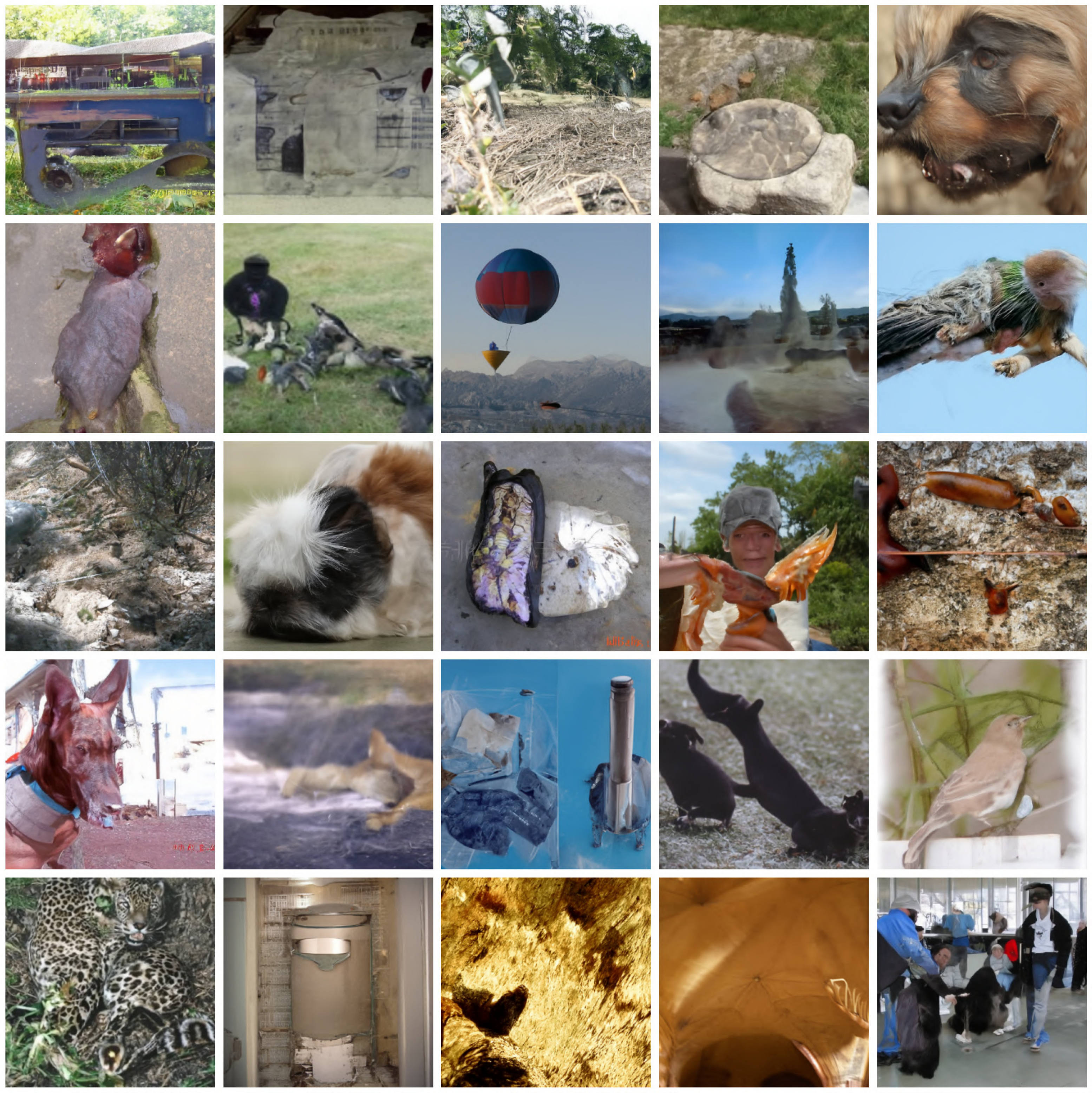}
    \caption{Random ImageNet samples generated with Classifier Guidance using $\lambda=1$.}
    \label{fig:cg_scale1}
    \includegraphics[width=.7\linewidth]{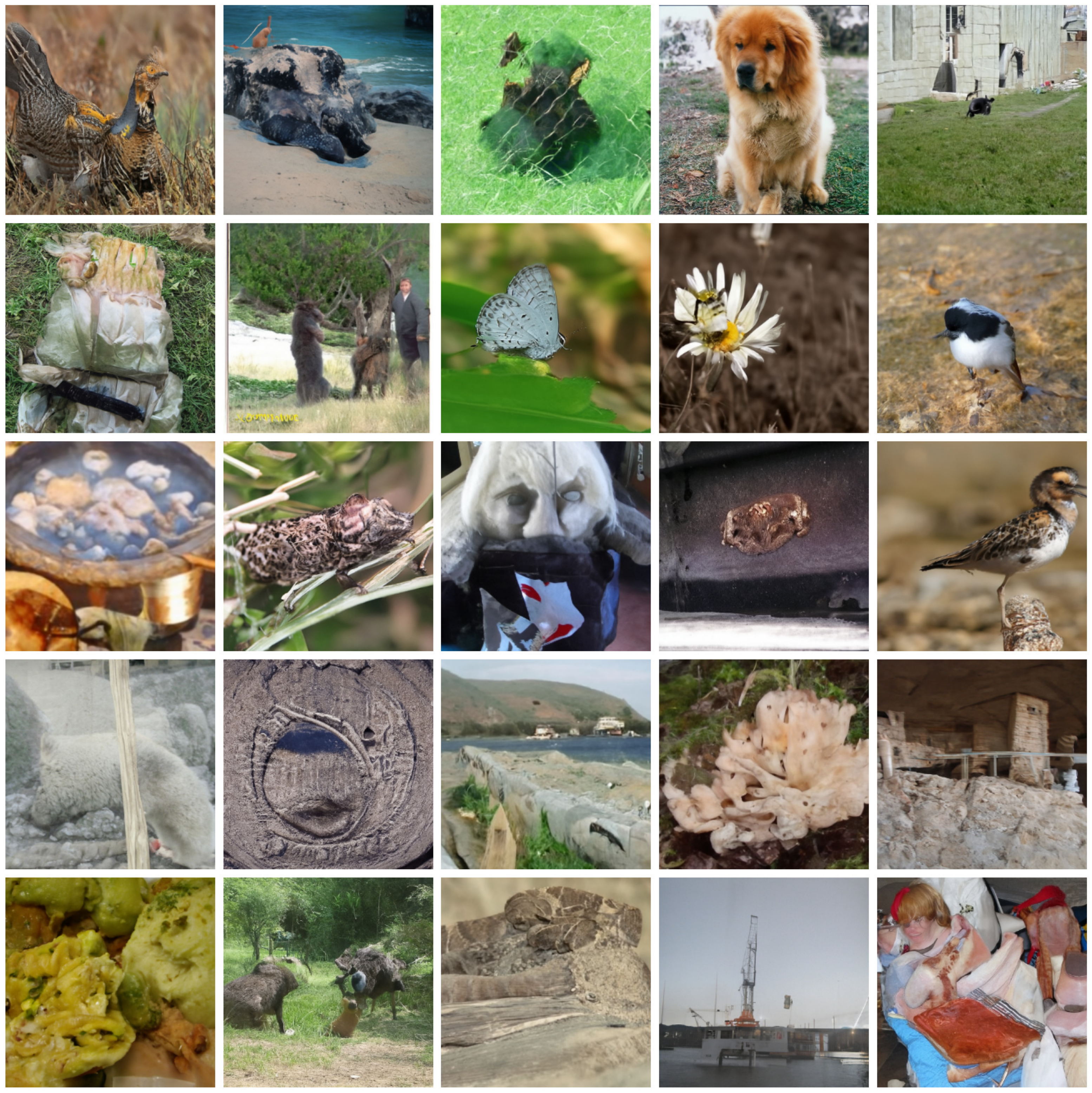}
    \caption{Random ImageNet samples generated with Dynamic Guidance using $\lambda=1$.}
    \label{fig:dg_scale1}
\end{figure}

\begin{figure}[t]
    \centering
    \includegraphics[width=0.7\linewidth]{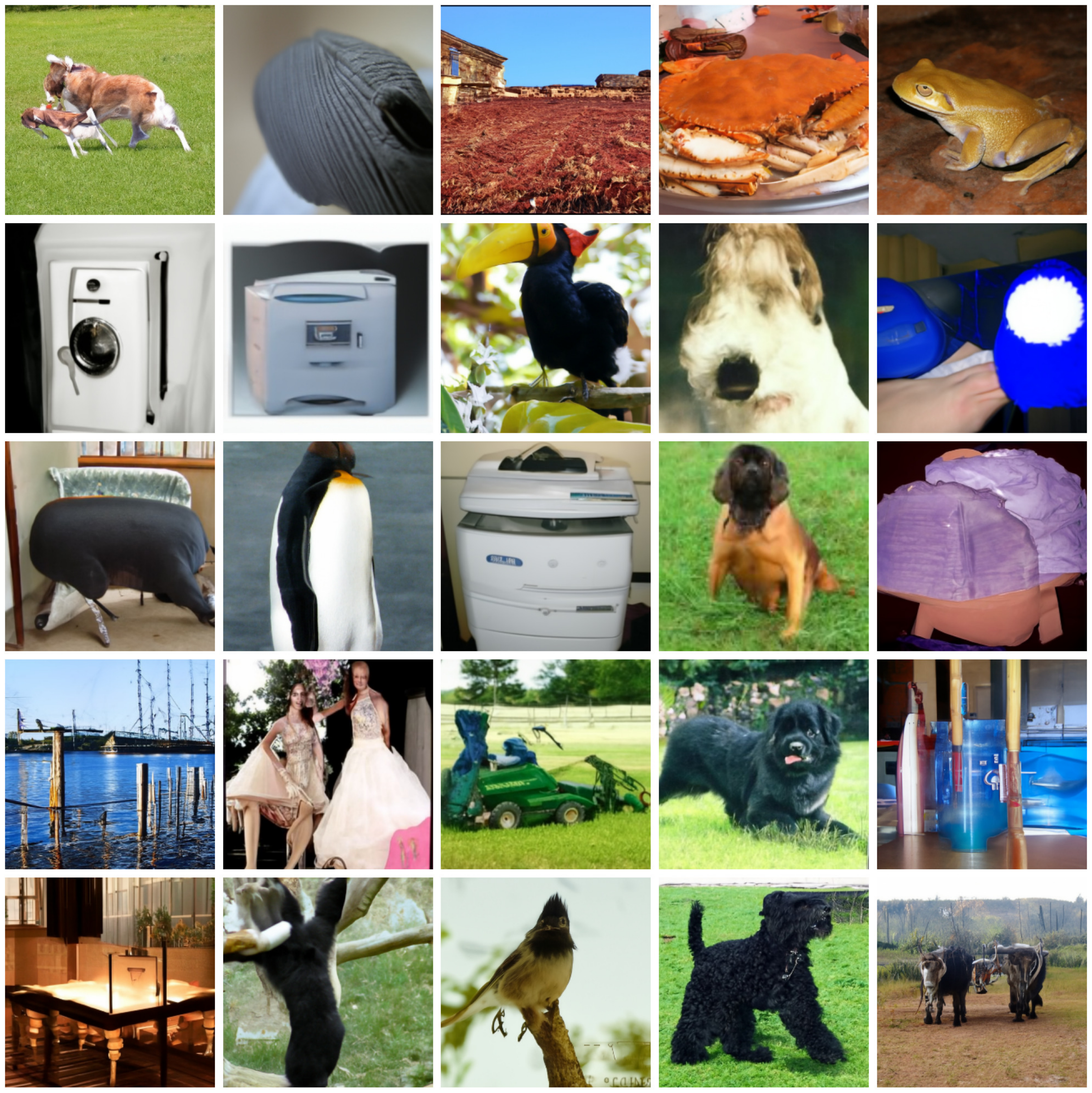}
    \caption{Random ImageNet samples generated with Classifier Guidance using $\lambda=10$.}
    \label{fig:cg_scale10}
    \includegraphics[width=0.7\linewidth]{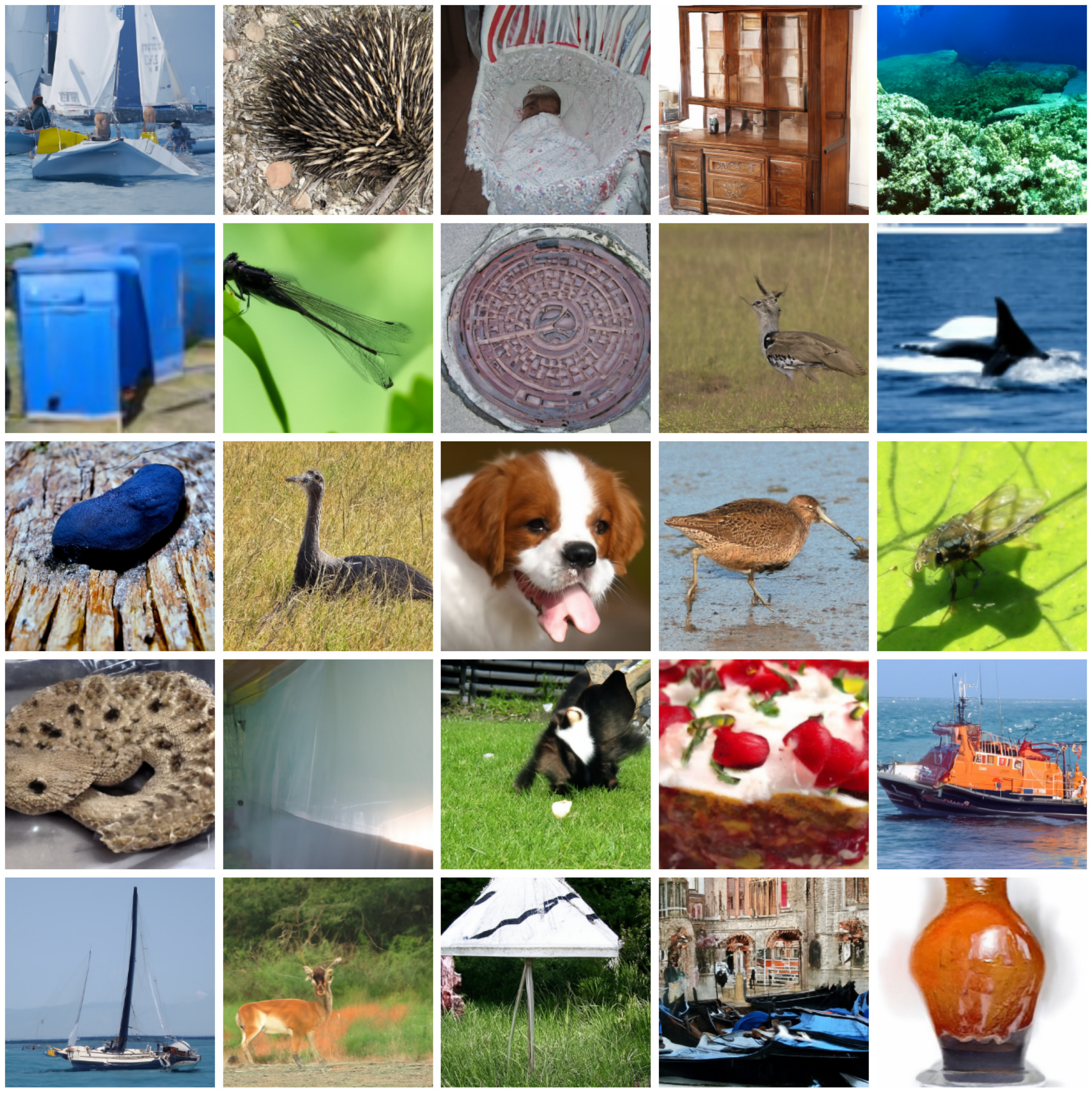}
    \caption{Random ImageNet samples generated with Dynamic Guidance using $\lambda=10$.}
    \label{fig:dg_scale10}
\end{figure}

\paragraph{Hands-11k.}
Figures \ref{fig:hands_ddim} and \ref{fig:hands_dmlcg} show 100 images of hands generated with DDIM and Dynamic Guidance, respectively.
\paragraph{ImageNet.}
Figures \ref{fig:cg_scale1}, \ref{fig:dg_scale1}, \ref{fig:cg_scale10}, and \ref{fig:dg_scale10} show examples of random images generated with Classifier or Dynamic Guidance with different guidance scales. Figures~\ref {fig:norm_llama} and \ref{fig:norm_random}, we show the difference in the norms of the denoising steps when using Classifier or Dynamic Guidance.

\begin{figure}[t]
    \centering
    \includegraphics[width=0.7\linewidth]{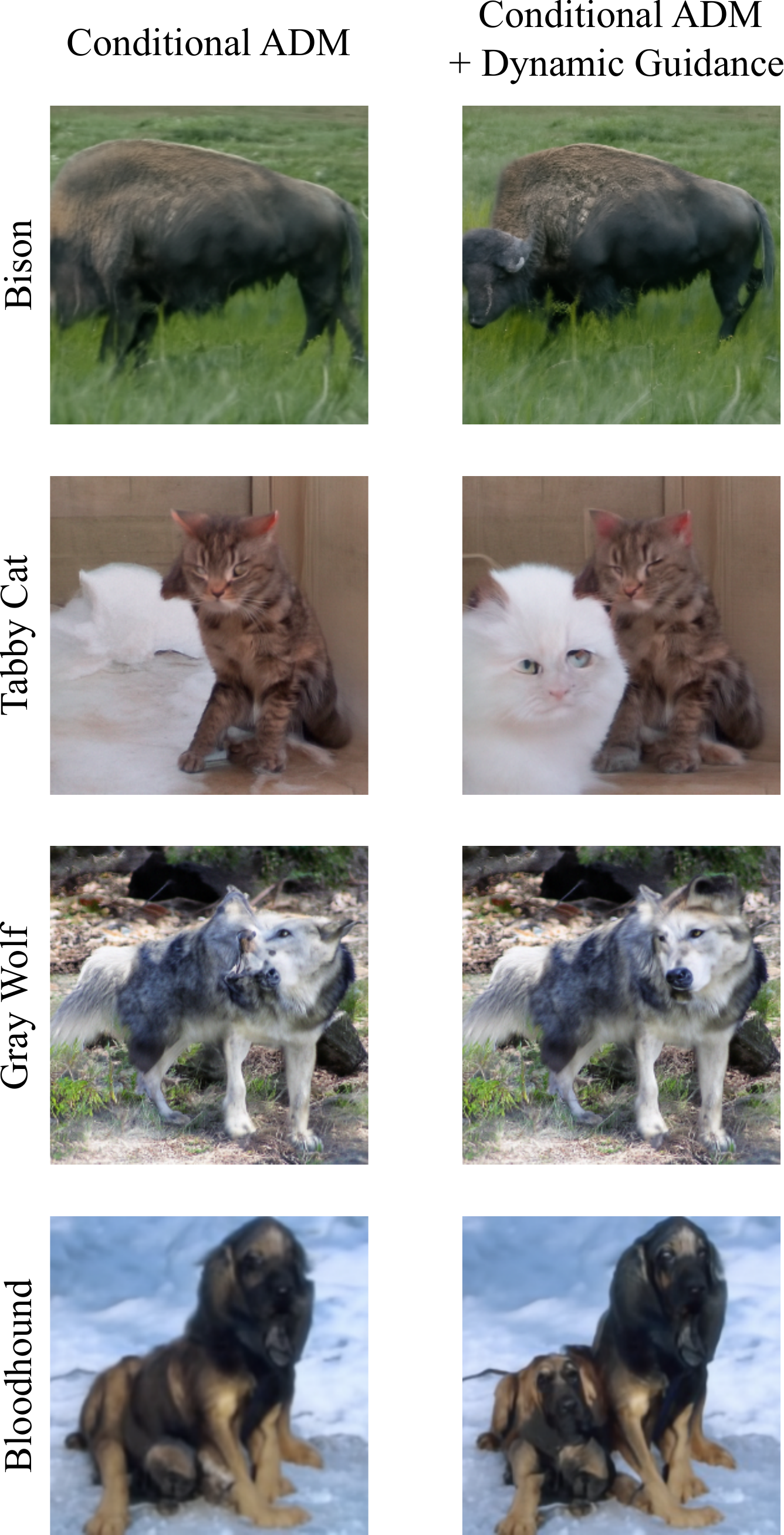}
    \caption{Samples generated from an ImageNet class-conditional ADM model, without guidance, and with Dynamic Guidance using the classifier trained on pseudoclasses created by clustering with DINOv2 embeddings (Section \ref{sec:dino}). We show that the conditioning and guidance labels do not need to be the same; Dynamic Guidance can improve generations even though the diffusion and guidance signals are not conditioned on the same classes. We practically see that the two can act orthogonally, with class-conditioning guiding the sample towards specific class-related attributes and and dynamic guidance helping avoid bad generations within the class. This is especially apparent in the second and last examples where Dynamic Guidance prevents hallucinations (cat with no face, dog with 5 visible legs) by guiding the samples towards clusters that represent multiple animals in an image.}
    \label{fig:cond}
\end{figure}

\paragraph{Class-Conditional ImageNet with Pseudo-Class Guidance.}
In Figure~\ref{fig:cond}, we show that Dynamic Guidance can be applied to a conditional model and improve generations even though the diffusion and guidance signals are not conditioned on the same classes.

\begin{figure}[t!]
    \centering
    \includegraphics[width=\linewidth]{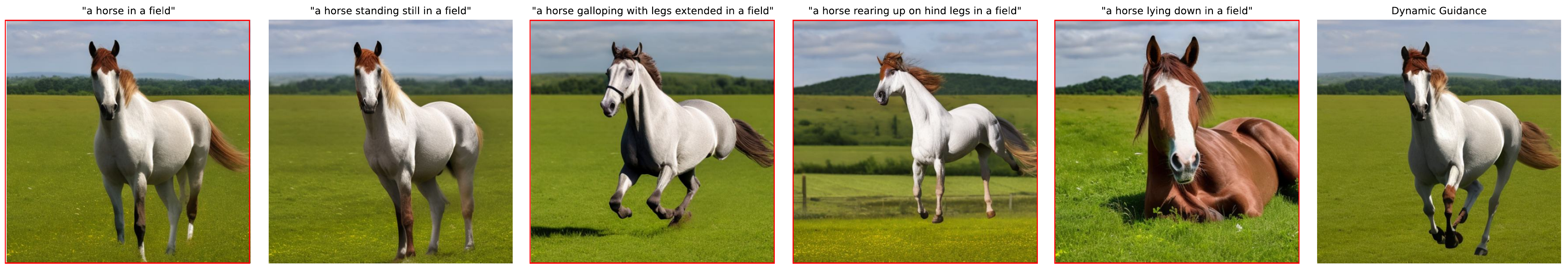}
    \caption{Using the generated, more complex, branches as base prompts can lead to worse generations and new hallucinations.}
    \label{fig:t2i_prompt}
\end{figure}

\begin{figure}[h]
    \centering
    \includegraphics[width=.8\linewidth]{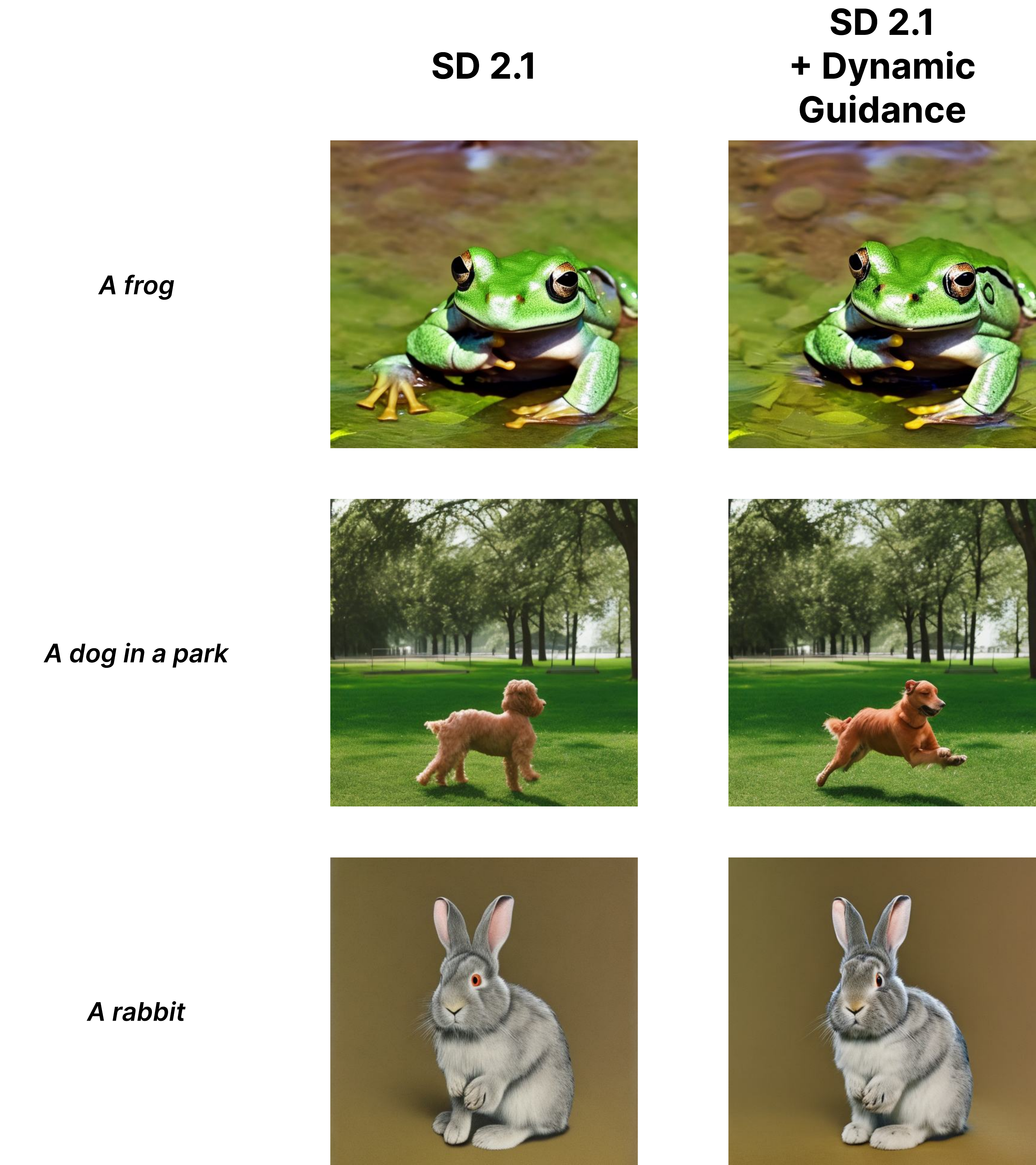}
    \caption{Examples of hallucinations in text-to-image generation that are fixed with Dynamic Guidance. In the three images generated by the baseline, the animals have an extra, hallucinated limb that is removed with Dynamic Guidance.}
    \label{fig:qual_t2i}
\end{figure}

\paragraph{Text-to-Image.} In Figures~\ref{fig:t2i_prompt} and \ref{fig:qual_t2i} we show examples of hallucinations in text-to-image generation that are fixed with Dynamic Guidance. 

\clearpage

\clearpage
\section{Potential Societal Impact}
\label{sec:societal_impact}
This work studies how to controllably mitigate hallucinations in diffusion models. Currently, people have been using common forms of hallucinations such as incorrect anatomy in generated pictures of humans or animals to detect and identify synthetically generated content. This means that developing methods that improve diffusion models by mitigating such hallucinations can potentially make identifying synthetically generated content such as deepfakes significantly more challenging for humans, even if automated detection is unaffected.

However, we contend that, as models naturally improve through increased scale and data diversity there is going to be a decrease in generated artifacts that are common across different models and very easily identifiable such as poorly generated hands, even though hallucinations will still be present. By understanding how to controllably mitigate hallucinations, our work facilitates the responsible and effective use of diffusion models in practical scenarios. Additionally, this evolution shows the requirement for additional future work on watermarking synthetically generated content so that it can be identified beyond the reliance on common visible failure modes of image generators.

\end{document}